%% file: main.tex
\documentclass[10pt,journal,compsoc]{IEEEtran}
\ifCLASSOPTIONcompsoc
  \usepackage[nocompress]{cite}
\else
  \usepackage{cite}
\fi

\ifCLASSINFOpdf
   \usepackage[pdftex]{graphicx}
\else
\fi
\usepackage{amsmath}
\usepackage{array}

\ifCLASSOPTIONcompsoc
 \usepackage[caption=false,font=footnotesize,labelfont=sf,textfont=sf]{subfig}
\else
 \usepackage[caption=false,font=footnotesize]{subfig}
\fi

\usepackage{times}
\usepackage{epsfig}
\usepackage{graphicx}
\usepackage{amsmath}
\usepackage{amssymb}
\usepackage{float}
\usepackage{bbm}
\usepackage{color}
\usepackage{mathtools}
\usepackage{booktabs}       
\usepackage{tabularx}
\usepackage{hhline}
\usepackage{multirow}
\usepackage{microtype}      
\usepackage[table,x11names,dvipsnames,table]{xcolor}
\usepackage{rotating} 
\usepackage{hyperref}

\DeclareMathOperator*{\E}{\mathbb{E}}

\newcommand\heading[1]{\bigskip\noindent\textbf{#1}}

\newcommand\edited[1]{#1}
\newcommand\wasanswered[1]{#1}

\definecolor{plotblue}{RGB}{101,158,213}
\definecolor{plotred}{RGB}{214,103,120}

\usepackage{array}

\newcolumntype{L}{>{}l<{}}
\newcolumntype{C}{>{}c<{}}
\newcolumntype{R}{>{}r<{}}

\usepackage{algorithm}
\usepackage{algpseudocode}
\algrenewcommand\algorithmicrequire{\textbf{Input:}}
\algrenewcommand\algorithmicensure{\textbf{Output:}}

\begin{document}
%
\title
{Improving Deep Metric Learning by \\Divide and Conquer}
%
%
%
%

\author{Artsiom~Sanakoyeu,
        Pingchuan~Ma,
        Vadim~Tschernezki,
        and~Bj\"{o}rn~Ommer
\IEEEcompsocitemizethanks{\IEEEcompsocthanksitem A. Sanakoyeu is the corresponding author.
\IEEEcompsocthanksitem All authors are with Heidelberg Collaboratory for Image Processing and Interdisciplinary Center for Scientific Computing, Heidelberg University, Germany. E-mail: {firstname.surname}@iwr.uni-heidelberg.de

\IEEEcompsocthanksitem Source code will be available at  \url{https://bit.ly/divconq-improved}.}
}

%
%

\markboth{TPAMI SUBMISSION}%
{}
\IEEEtitleabstractindextext{%
\begin{abstract}
Deep metric learning (DML) is a cornerstone of many computer vision
applications. It aims at learning a mapping from the input domain to an embedding space, where semantically similar objects are located nearby and dissimilar objects far from another.
The target similarity on the training data is defined by user in form of ground-truth class labels.
However, while the embedding space learns to mimic the user-provided similarity on the training data,
it should also generalize to novel categories not seen during training.
Besides user-provided groundtruth training labels, a lot of additional visual factors (such as viewpoint changes or shape peculiarities) exist and imply different notions of similarity between objects, affecting the generalization on the images unseen during training.
However, existing approaches usually directly learn a single embedding space on all available training data, struggling to encode all different types of relationships, and do not generalize well.
We propose to build a more expressive representation by jointly splitting the embedding space and the data hierarchically into smaller sub-parts.
We successively focus on smaller subsets of the training data, reducing its variance and learning a different embedding subspace for each data subset. Moreover, the subspaces are learned jointly to cover not only the intricacies, but the breadth of the data as well.
Only after that, we build the final embedding from the subspaces in the conquering stage. The proposed algorithm acts as a transparent wrapper that can be placed around arbitrary existing DML methods. Our approach significantly improves upon the state-of-the-art on image retrieval, clustering, and re-identification tasks evaluated using CUB200-2011, CARS196, Stanford Online Products, In-shop Clothes, and PKU VehicleID datasets.

\end{abstract}

}

\maketitle

\IEEEdisplaynontitleabstractindextext

%
\IEEEpeerreviewmaketitle


%



\IEEEraisesectionheading{\section{Introduction}\label{sec:introduction}}

\IEEEPARstart{D}{eep} metric learning methods learn to measure similarities or distances between arbitrary pairs of data points, which is of paramount importance for a number of computer vision and machine learning applications. Deep metric learning has been successfully applied to image search \cite{bell2015learning_contrastive,huang2016unsupervised,lifted_struct,deepranking2014}, person/vehicle re-identification \cite{chopra2005learning,vehicleid,hdc,roth2019mic}, fine-grained retrieval \cite{proxynca}, near duplicate detection \cite{zheng2016improving}, clustering \cite{deep_clustering2016} and zero-shot learning \cite{lifted_struct,hist_loss,cliquecnn2016,sanakoyeu2018pr,buchler2018improving}. The exponential increase of easily accessible digital images and the importance of image retrieval and related tasks for numerous applications call for fast and scalable deep metric learning approaches.


\begin{figure}[t]
\begin{center}
 \includegraphics[width=1.0\linewidth]{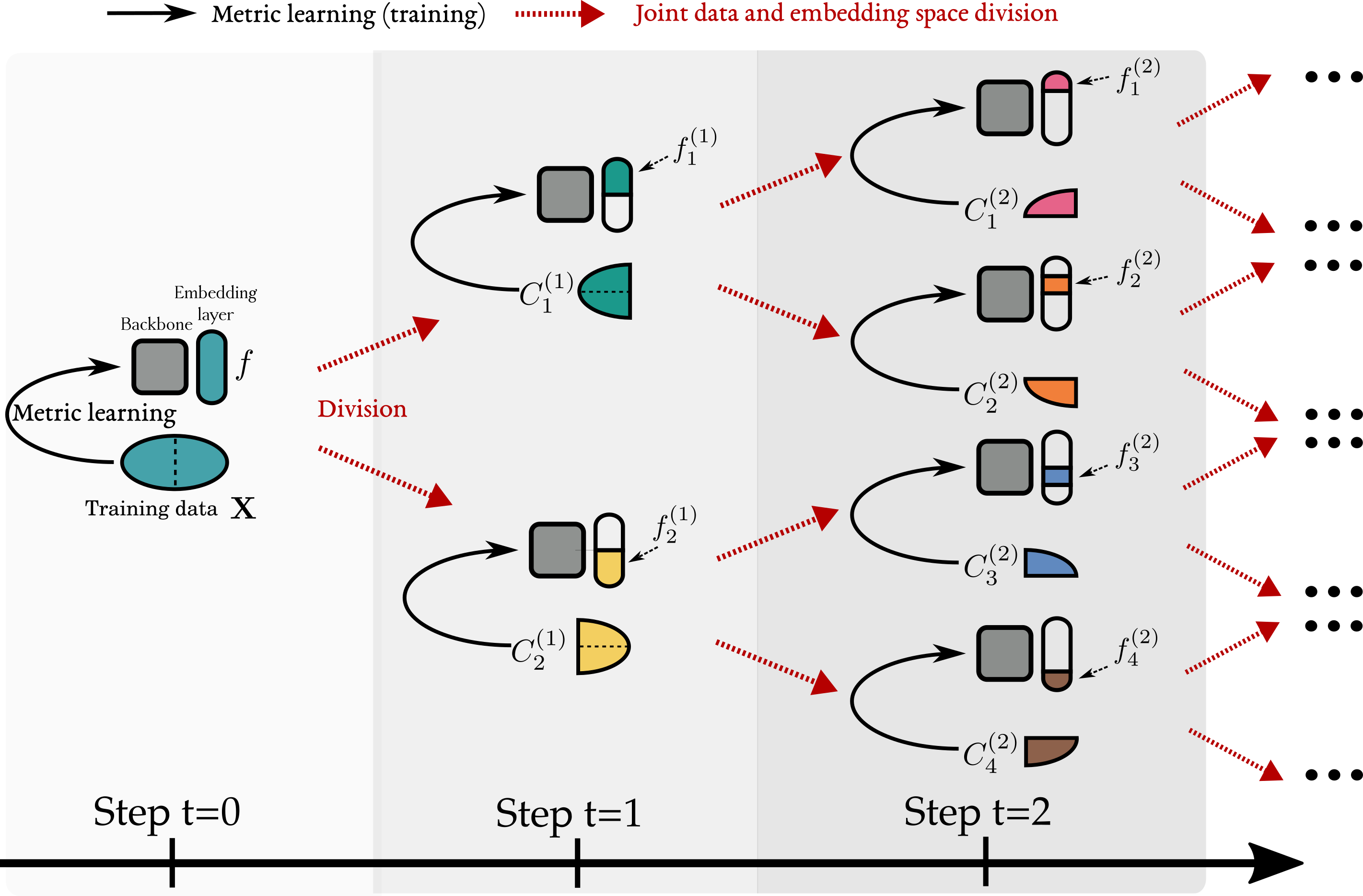}
\end{center}
   \caption{The pipeline of the proposed approach. We simultaneously split the training data and the embedding space at every division step $t$.
   At every $t$ number of sub-problems, which are optimized jointly, increases. $C_k^{(t)}$ denotes the cluster of images belonging to sub-problem $k$ and $f_k^{(t)}$ is the subspace of the embedding space assigned to it. All network parameters are shared across all steps $t$.
   }
\label{fig:pipeline}
\end{figure}

Deep metric learning methods learn a mapping from an input domain such as RGB images to an embedding space with a simple fixed distance metric which implicitly captures the similarity between the inputs.
To learn such an embedding space, ground-truth classes are typically used as a proxy for the target image similarity: Two images are considered similar if they belong to the same ground-truth class and the DML loss functions are designed in such a way that similar images are mapped close in the learned embedding space and dissimilar images far away.
DML aims at solving a transfer problem, i.e. the typical scenario is to train on one set of categories and evaluate on a completely different set of test categories.
Therefore, the main goal is to learn such an embedding space that is able to generalize to previously unseen images and categories.



The major challenge for DML approaches is that they are facing a great number of different latent factors which cause images to be similar or dissimilar,
where the ground truth classes are only one of many factors which contribute to the image similarity \cite{veit2017conditional,tan_saenko2019unsup_sim_notions}.
Image similarity can be based on different semantic or visual attributes of the images \cite{su2015_attributes_for_person_reid,gan2016learning_attributes}. Moreover, different images may feature different attributes with none being present in all. Consequently, there is a large intra-class variability, which aggravates directly learning a single representation and metric of similarity for all the data. For example, two images of cars can be similar based on such factors as manufacturer, model, car body style (sedan, wagon, pickup, etc.), body color, number of wheels, viewpoint, geolocation, background, weather conditions and many more.
Moreover, images can show similarities based on accidental traits such as compression artifacts, noise patterns and image distortions exhibited by photo-sensors \cite{lukas2006digital,tuama2016camera}.
However, existing deep metric learning approaches strive to directly learn a single embedding space for all samples from the training data distribution \cite{facenet,npairs,margin,wang2019multisim_loss}.
As a result, they (i) cannot explain all of the aforementioned latent factors directly and capture only those factors which are important for distinguishing merely the training classes.
For example, if we want to discriminate bird species and a lot of hawk images have sky in the background and most of the falcon images have trees in the background, then the presence of trees in the background will be regarded as a very important feature by the network. However, it does not capture any attribute of a bird, like, for instance, bird head shape, and would not generalize to novel test categories where the background is not informative.
Moreover,
(ii) the neural networks can rely on some visual traits (e.g. compression artifacts, noise patterns, image distortions, etc.) as ``shortcuts" to overfit the training distribution \cite{noroozi2016unsupervised}.
These shortcomings cause poor generalization of the learned representations on previously unseen test images.
To address these issues and enforce more structure (not captured by the ground truth labels) in the embedding space we need to
(i) decrease the influence of misleading visual factors
by grouping the data according to the latent factors into more homogeneous subsets and
(ii) constrain the embedding space by enforcing different subspaces to focus on specific data subsets featuring different latent attributes.
Hence, we propose to explicitly train different embedding subspaces on
different subsets of the data where data subsets are constructed to reduce
the variability and the variation of the latent factors in the embedding space.
Returning to our previous example, it would allow us to group all falcons and hawks with trees in the background and enforce an embedding subspace to ignore the background and look for more informative visual attributes.


We propose a novel deep metric learning approach which decomposes the original problem into a set of smaller ones, inspired by the well-known ``divide and conquer" strategy.
It is hard to capture the breadth and the details of all the data at once, therefore
we progressively partition the original dataset into more homogeneous subsets while simultaneously
subdividing the embedding space into non-disjoint embedding subspaces for each subset and jointly optimizing them.
We (i) explicitly minimize the variance of the training data by dividing it into subsets via clustering, based on the similarity learned so far. Such a subdivision results in implicit mining of harder training samples, because the average pairwise distance between samples within clusters is smaller than in the entire dataset.
And (ii) we learn specific embedding subspaces for each subset of the data.
By focusing on smaller, more homogeneous sets of data, small specific embedding spaces for each subset are forced to measure similarities based only on the common attributes discovered in each subset. Together the smaller embedding spaces exhibit better
%
generalization to novel classes.
Applying our ``divide and conquer" strategy recursively to yield a hierarchical decomposition allows to increase the granularity of learned relationships over training iterations by successively zooming in and focusing on more subtle relationships rather than trying to reconcile all of them in one shot.
All individual subspaces are optimized jointly, covering the breadth of the training data.
The individual subspaces are smaller than the overall embedding space so that together their size does not increase
the overall dimensionality of the full embedding.
Lastly, in the conquering stage, we construct the full embedding by combining individual embedding subspaces together. Fig.~\ref{fig:pipeline} presents a schematic pipeline of our approach.

Our method addresses issues of existing metric learning while acting as a transparent wrapper that can be placed around arbitrary existing DML approaches. Because of the hierarchical nature of our approach we show that we can even reduce an overfitting commonly happening  when training embeddings of very large dimensionality ($1024$ or $2048$) \cite{lifted_struct,roth2019mic}.
We demonstrate the ability of the proposed approach to learn fine-grained image similarities, achieving state-of-the-art retrieval and re-identification performance on five benchmark datasets for fine-grained similarity learning: CUB200-2011 \cite{cub200_2011}, CARS196 \cite{cars196}, Stanford Online Products \cite{sop}, In-shop Clothes \cite{deepfashion}, and PKU VehicleID \cite{vehicleid}.


\section{Related work}

Metric learning has for long been of major interest for the vision community, due to its broad applications including object retrieval \cite{lifted_struct,hist_loss,desc1}, zero-shot and single-shot learning \cite{hist_loss,lifted_struct}, keypoint descriptor learning \cite{desc_learn}, face verification \cite{chopra2005learning,facenet,center_loss}, vehicle identification \cite{vehicleid,chu2019vehicle_view_aware}, visualization of high-dimensional data \cite{hadsell2006contrastive,maaten2008tsne}, and clustering \cite{deep_clustering2016}.

\heading{Loss functions.}
Contrastive loss \cite{hadsell2006contrastive,bell2015learning_contrastive} is one of the simplest metric learning losses which pulls together pairs of samples with the same class label and pushes apart pairs of samples from different classes.
Recently, a lot of research efforts have been devoted to designing new loss functions \cite{facility_loc,hist_loss,npairs,angular,proxynca,wang2019multisim_loss}.
For example, Facility Location \cite{facility_loc} optimizes a clustering quality metric, Histogram loss \cite{hist_loss} minimizes the overlap between the distribution of positive and negative distances, Angular loss \cite{angular} imposes extra geometrical constraints in the embedding space. \cite{lifted_struct,npairs} introduce a soft formulation of the triplet loss \cite{weinberger2009triplet} replacing hinge function with NCA~\cite{nca} which does not require tuning of a margin parameter. Recently, Yu et al.~\cite{tuplet_margin_loss2019} modified N-Pairs loss by introducing a margin and a temperature scaling in the objective, and an extra loss term which penalizes high intra- and inter-class pairwise distance variance.
Proxy-based losses \cite{proxynca,kim2020proxy_anchor,EDMS_2019_CVPR} further extend the NCA paradigm by computing proxies (prototypes) for the training classes in the dataset and optimizing the distances to these proxies using the NCA objective\cite{nca}. Proxy-based losses are closely related to the classification-based deep metric learning methods \cite{liu2016large_margin_softmax,zhai2018making_normsoftmax,wang2018cosface}. In this case, the training images are classified using softmax function where the columns of the weight matrix of the classification layer represent the prototypes for the classes.
Magnet loss \cite{magnet} is similar to proxy-based losses, but it does not learn the class prototypes, instead it splits every ground truth class in sub-classes by clustering and pushes training samples closer to the precomputed centroids of the corresponding sub-classes.
Wen et al.~\cite{center_loss} proposed to use a center loss jointly with the softmax classification loss to enforce more clustered representations by minimizing the Euclidean distance between the image embeddings and the learnable centroids of the corresponding classes.
MIC~\cite{roth2019mic} models visual characteristics shared across classes by utilizing an extra surrogate loss discriminating between data clusters.
SoftTriple Loss \cite{qian2019SoftTriple} upgrades the softmax classification loss by learning multiple prototypes for each class allowing to capture several modes for the classes with high intra-class variance.
The major drawback of proxy- and classification-based losses is the limited scalability with respect to the number of classes.
Another type of loss is FastAP \cite{cakir2019fastAP} which aims at ranking the images by optimizing the non-differentiable Average Precision (AP) measure. The authors use a probablistic interpretation of the AP and approximate it by distance quantization and histogram binning.

Our work is orthogonal to all aforementioned approaches, as it provides a framework for learning a distance metric that is independent of the choice of a particular loss function.

\heading{Informative sample mining.}
It is common for metric learning methods to use pairs \cite{ConvNetSimPatch} or triplets \cite{weinberger2009triplet,ConvNetSimTriplet,facenet} of samples. Some even use quadruplets \cite{hist_loss} or impose constraints on tuples of larger sizes \cite{lifted_struct,npairs,posets,hermans2017in_defense_triplet_loss}.
Using tuples of images as training samples yields a huge amount of training data. However, only a small portion of the samples among all $N^p$ possible tuples of size $p$ is informative because most of the tuples produce very low loss and insufficient gradient for learning.
A lot of works attempted to explore how to select the most informative tuples 
for training, and such approaches can be categorized in two groups: (a) methods which focus on meaningful sample mining within a randomly drawn mini-batch, and (b) the methods which mine the entire dataset but require computationally expensive preprocessing steps.

The first group of methods strive to find informative samples within a randomly drawn mini-batch (\emph{local mining}).
Some of the methods from this category utilize all pairwise relationships within a mini-batch \cite{lifted_struct,ding2015person_reid_batch_all,tadmor_multibatch_2016}, others mine hard negative \cite{facenet,hermans2017in_defense_triplet_loss} or easy positive pairs \cite{xuan2020easy_positive}. Wang et al.~\cite{rankedlist2019} consider all possible pairs within the batch but set the weights for negative pairs as the exponent of the margin violation magnitude.
MS-loss~\cite{wang2019multisim_loss} generalizes tuple-based losses and reformulates them as different weighting strategies of positive and negative pairs within a mini-batch. Wu et al.~\cite{margin} sample negative examples uniformly according to their relative distance to the anchor, and, recently, Roth et al.~\cite{roth2020pads} proposed to learn the distribution for sampling negative examples instead of using a predefined one.
 DAML~\cite{daml} generates synthetic hard negatives for the current mini-batch using an adversarial training.
 HTG \cite{hard_triplet_gen} also employs adversarial training to alter a given triplet by pushing embedding vectors of the images from the same class apart while pulling embedding vectors of the images from different classes closer.
DVML~\cite{dvml} assumes that the distribution of intra-class variance is independent on the class label and makes a negative example harder by adding a variance component sampled from the learned distribution.
Another approach that synthesizes hard negatives is HDML \cite{hdml}, it uses linear interpolation to move the negative example closer to the positive one in the embedding space, thus increasing the triplet hardness.
The drawback of the \emph{local mining} methods
is the lack of global information while having only a local view of the data based on a single randomly-drawn mini-batch of images. As a consequence, the performance of such approaches strongly depend on the mini-batch size, which is limited by the GPU memory size.

The second group of mining approaches have a global view of the data and utilize the entire dataset for finding samples which provide the largest training signal (\emph{global mining}).
Wang et al.~\cite{deepranking2014} relies on a highly optimized handcrafted ``golden feature" to compute the pairwise and unary relevance scores for all images in the dataset. The closer the image to the class centroid, the higher unary relevance it gets. The anchor, positive and negative examples are then sampled according to the relevance scores. However, the ``golden feature" is expensive to compute, difficult to develop, and is crafted for a specific dataset.
\wasanswered{Later works \cite{smart_mining,htl,roth2020revisiting,iscen2018mining} do not rely on hand-crafted features and mine hard negative examples in the learned embedding space.
However, these sampling techniques require either running an expensive preprocessing step (quadratic in the number of data points) for the entire dataset and for every epoch \cite{smart_mining,htl,roth2020revisiting,iscen2018mining} or utilize additional meta-class labels \cite{cakir2019fastAP} for hard sample mining.}
Suh et al.~\cite{suh2019softmax_and_triplet} sample negative pairs of images from the most confused classes. But this approach requires a joint training of the classification head using softmax loss and the embedding layer using a triplet loss.
On the contrary, our approach efficiently alleviates the problem of the abundance of easy samples by
jointly dividing the data and the embedding space without an extra classification head. The data is clustered  in the embedding space learned so far, so samples drawn from the clusters have smaller pairwise distances than those which are randomly drawn from the entire dataset. This serves as a proxy for informative sample mining in the entire dataset, and it is computationally efficient due to a linear time K-means implementation by \cite{faiss}. \wasanswered{Also, in contrast to \cite{htl,cakir2019fastAP}, our method does not rely on ground truth class or meta-class labels for building the data hierarchy.}
Moreover, our approach is complementary to within-batch mining methods and can be jointly used with them.

\begin{figure}[t]
\begin{center}
\includegraphics[width=\linewidth]{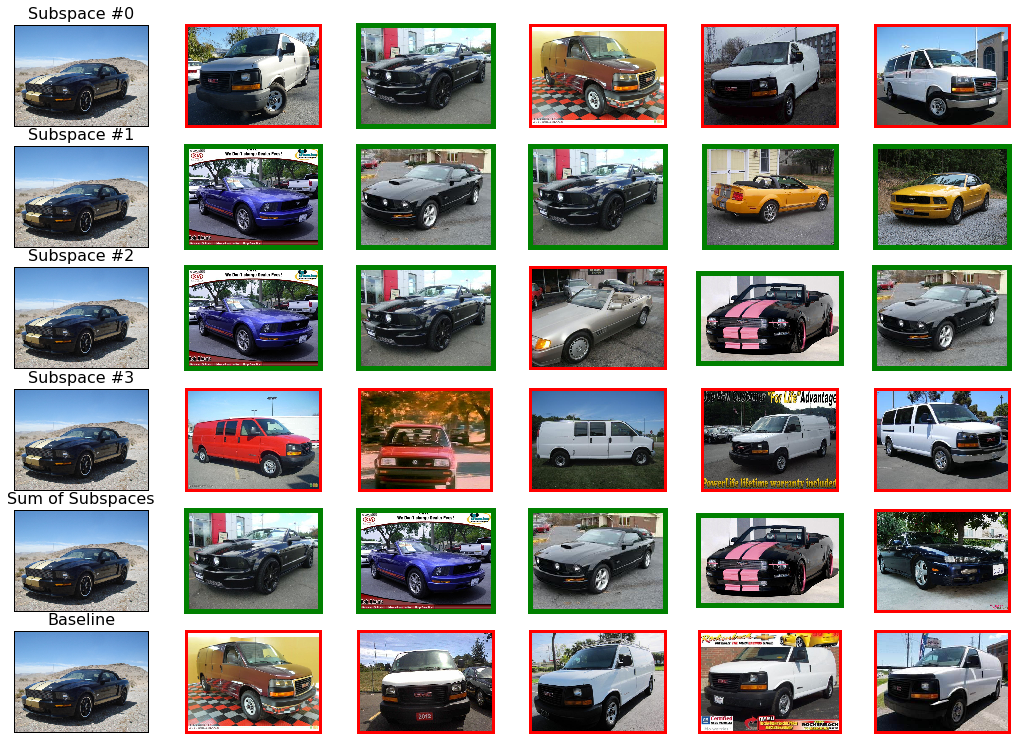}

\end{center}
   \caption{Comparison of the retrieval results on CARS for our approach with 4 subspaces and the baseline approach (no division in subspaces). First $4$ rows show retrieved nearest neighbors using different subspaces. The 5-th row: the results for the sum of all subspaces. Green border: true positive, red border: false positive. The queries and retrieved images are taken from the test set. Subspace \#3 makes similar mistakes as the baseline model, but our subspaces are able to correct each other which yields more accurate results when all subspaces are combined together.}
\label{fig:cmp_learners_vs_baseline_cars}
\end{figure}

\heading{Ensembling for deep metric learning.}
Another line of work in deep metric learning is ensemble learning \cite{hdc,a_bier,adaboost,cnn_ensemble_detection}.
Previous works \cite{hdc,a_bier} employ a sequence of ``learners" with increasing complexity and mine samples of different complexity levels for the next ``learners" using the outputs of the previous learners.
HDC \cite{hdc} uses a cascade of multiple models of a specific architecture and trains earlier layers of the cascade with easier examples while harder examples are harnessed in later layers. A-BIER \cite{a_bier} applies a gradient boosting learning algorithm to train several learners inside a single network in combination with an adversarial loss \cite{ganin2016domain,goodfellow2014generative}.
ABE \cite{kim2018_abe} introduces soft attention in the intermediate convolutional feature maps to focus on different parts of the objects combined with extra divergence loss to diversify the attention.
DREML~\cite{dreml} and EDMS~\cite{EDMS_2019_CVPR} train multiple networks on random splits of the data using variants of the ProxyNCA~\cite{proxynca} loss. The downside of these approaches is the drastic increase of the number of parameters and the computational cost.
The key difference of the aforementioned approaches to ours is that we do not have an ensemble of the networks, but train a \emph{single} network by splitting the embedding space and clustering the data jointly, so each ``learner" is assigned to the specific subspace and corresponding portion of the data. The ``learners" are jointly trained on non-overlapping chunks of the data which reduces the training complexity of each individual ``learner", facilitates the learning of decorrelated representations and can be easily parallelized.
Moreover, our approach does not introduce extra parameters during training since we do not alter the architecture and utilize only a single network. It does not require any elaborate loss functions but can be applied to arbitrary losses and any existing network architecture.

\heading{Curriculum learning.}
Curriculum learning describes a strategy which defines the order of the learning tasks and was first introduced in the seminal work \cite{bengio2009curriculum}.
Gradually increasing the complexity of the concepts and showing easier examples before more difficult ones is the idea which naturally arises from observing how animals and humans learn \cite{krueger2009flexible_shaping,bengio2009curriculum}.
 The optimal order can significantly improve the convergence speed and model generalization \cite{bengio2009curriculum}.
The idea of increasing task difficulty while learning has been utilized in different applications. Examples are in robotics \cite{sanger1994robot_increase_difficulty}, training of deep networks \cite{hinton2006fastdeepbelieve,wang2017deepgrowing,karras2018progressive}, reinforcement learning \cite{matiisen2019teachercurriculum}, meta-learning \cite{sun2019meta_curriculum}, and time series forecasting~\cite{koenecke2019curriculum_financial} to name just a few.
Informative sample mining, discussed earlier in this section, also fits under the umbrella of curriculum learning. However, curriculum learning is a significantly broader framework which includes the progressive increase of the training setup complexity over training iterations.
In our approach, we also employ a curriculum. To achieve a smooth transition from an easy towards a more complex training setup, we progressively increase the number of learning subtasks (i.e. clusters and subspaces) during training.

\heading{Local and hierarchical metric learning.}
Local metric learning methods (LML) \cite{saxena2015coordinated_LML,amand2017sparse_comp_LML,bohne2018pairwise_LML} learn a collection of Mahalanobis distance metrics (which is equivalent to learning a collection of linear projections) each operating on a different subset of the data obtained by K-means or Gaussian mixture clustering. During inference with LML methods, test samples are mapped in the final embedding space computed as a linear combination of the local projections.
In similar spirit, we learn specific embedding subspaces for each subset of the data. However, our approach is more general and flexible. In our case the learned mappings to the subspaces are non-linear, they are parametrized by convolutional neural networks (CNNs) and operate on raw RGB pixels instead of fixed precomputed features, and the subspaces do not have a fixed dimensionality as they are learnable. Moreover, our recursively ”divide and conquer” strategy yields a hierarchical decomposition of the data and the embedding space which allows to progressively increase the granularity
of learned relationships over training iterations.

Works \cite{yan2015hd,caron2019unsupervised} learn \edited{a two-level category hierarchy by using coarse and fine classifiers}. However, these approaches are limited to classification and do not explicitly learn a ranking in the embedding space. HTL~\cite{htl} builds a data hierarchy for metric learning, which requires ground truth classes and an expensive (quadratic in the number of data points) preprocessing step every epoch. Moreover, in contrast to our approach, \cite{htl} does not learn multiple subspaces.






\begin{figure}[t]
\begin{center}
\includegraphics[width=\linewidth]{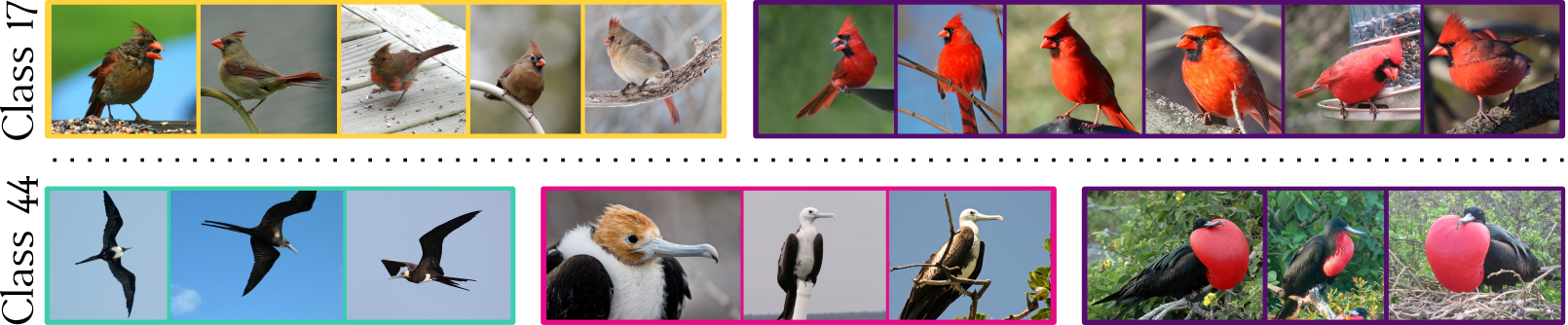}

\end{center}
   \caption{Visualization of the images sharing the same GT class label but splitted across several clusters by the proposed division procedure ($K_{max} = 4)$ for a model trained on CUB. Different border colors correspond to different clusters. We can see that the clustering splits the classes according to different visual modes. For example, for class $44$ (frigatebird): Female and juvenile frigatebirds have white chests while adult males are entirely black save for the the bright red throat pouch, which is also captured by our clustering.}
\label{fig:cub_classes_splitted}
\end{figure}

\section{Method}



The aim of DML is to learn a similarity measure between an arbitrary pair of images $x_i$ and $x_j$. 
During training, the target similarity is provided by a user in the form of discrete ground-truth classes, since it is infeasible to obtain continuous ground-truth similarity scores.
We then need to learn a mapping $f$ to an embedding space so that a predefined distance measure in this space captures the desired similarity between the images, i.e. the distances between images from the same class should be small and the distances between images from different classes should be large.
The mapping $f$ is learned directly from RGB values using a CNN which maps an RGB image $x_i$ onto a point $f(x_i)$ in the $d$-dimensional embedding space $\mathbb{R}^d$. The distance between two points $i$ and $j$ in the embedding space is then computed using a predefined distance measure, e.g. $D_{i,j}=||f(x_i) - f(x_j)||_2$.


Our approach is applicable irrespective of the particular DML loss function. Therefore, merely for simplicity of demonstration in this section we will use triplet loss \cite{weinberger2009triplet,facenet}, one of the most renown loss functions for deep metric learning.
Let us consider a tuple of images $(x_a, x_p, x_n)$, where $x_a$ is an anchor image, $x_p$ is a positive image from the same class as the anchor image and $x_n$ is a negative image from any other class. The triplet loss pushes the anchor image $x_a$ closer to the positive image $x_p$ and further from the negative image $x_n$ and is defined as
\begin{equation}\label{eq:triplet_loss}
    l_{\text{triplet}}(a,p,n)  = \left[ D_{a, p}^2 - D_{a, n}^2  + \alpha \right]_+,
\end{equation}
where $[\cdot]_+$ denotes the positive part
and $\alpha$ is the margin.

\subsection{Division}\label{sec:divide}
To start, we train the mapping $f$ using triplet loss on the entire dataset.
We strive to learn a similarity measure that can generalize well to differently distributed data from test classes unseen during training.
But some visual attributes can be misleading, despite high correlation with the ground truth labels, and cause overfitting to the training data.
To decrease the influence of misleading visual factors we propose to focus on smaller more homogeneous subsets (clusters) of the data with reduced variance and optimize only a specific subspace of the embedding space for each subset.
We further show that our approach results in (i) implicit mining of harder training samples and (ii) a more expressive embedding space with stronger generalization to novel classes due to learning several subspaces jointly, while focusing on different latent factors.
Such decomposition can be done by directly splitting the data and the embedding space in the desired number of subparts from the onset \cite{sanakoyeu_dcesml}. But many research works \cite{hinton2006fastdeepbelieve,bengio2009curriculum,wang2017deepgrowing,matiisen2019teachercurriculum,karras2018progressive} argue that a smooth transition from an easy training setup to a more complex one is usually a more fruitful strategy which improves training convergence and generalization. Thus, we progressively increase the number of subtasks (i.e. clusters and subspaces) during training while keeping the dimensionality of the embedding space fixed. We start the embedding space training on the entire dataset and during training iterations we progressively decompose the embedding space into subspaces and also the data into subsets (see Fig.~\ref{fig:pipeline}). This increases the granularity of learned relationships during training by gradually focusing on more details of smaller subsets of the data and the embedding space. However, we also cover the breadth of the data by training all individual subspaces jointly within a single neural network.


\begin{figure*}[t]
\begin{center}
\includegraphics[width=0.96\linewidth]{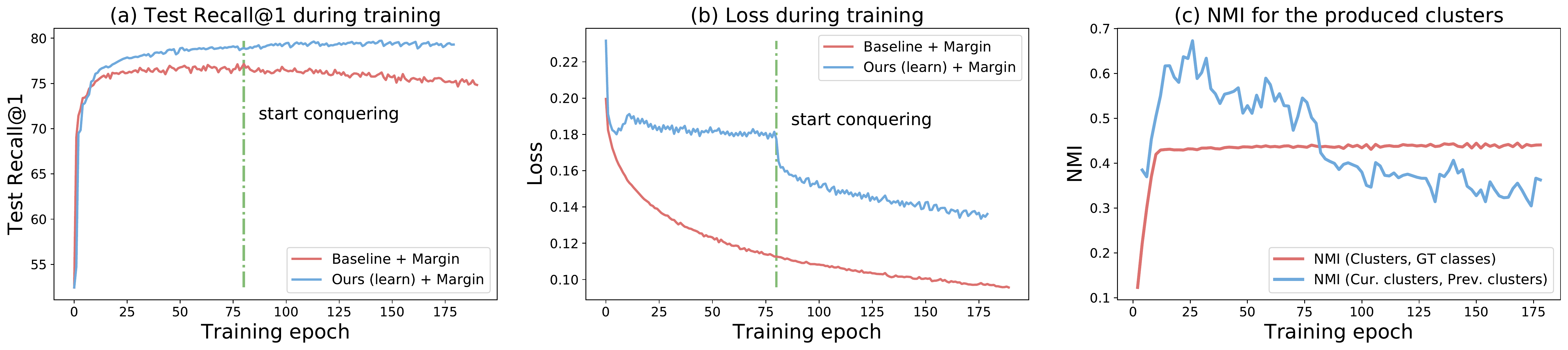}

\end{center}
   \caption{
   \textbf{(a)} Recall@1 and \textbf{(b)} loss for the baseline ($K_{max}=1$) and our proposed approach ($K_{max}=32$) during training on the SOP with margin loss \cite{margin}. \textbf{(c)} For our approach we show NMI between the produced clusters and the GT classes in \emph{red}, and NMI between clusters at the current and the previous division steps in \emph{blue}.
   }
\label{fig:plot_loss_nmi}
\end{figure*}

\subsubsection{Dividing the data and reducing the variance}\label{sec:reducing_the_variance}
We need to partition the training data  $\mathbf{X} = \{x_1, \dots, x_N \}$ in a way that minimizes the variance of each subset in the embedding space. This corresponds to minimum variance clustering, with K-means \cite{kmeans1967MacQueen} being one of the standard algorithms. It partitions the data into $K$ subsets $\mathbf{C} = \{C_1, C_2, \dots, C_K\}$, $\mathbf{X} = \bigcup_{k=1}^K C_k$, so that the variance of the resultant parts is explicitly minimized,
\begin{equation}\label{eq:kmeans}
\begin{aligned}
   \mathbf{C} =& \underset{\mathbf{C}} {\operatorname{argmin}}  \sum_{i=1}^{K} \sum_{x \in C_i} \left\| f(x) - \E[f(C_i)] \right\|^2 \\
   =& \underset{\mathbf{C}} {\operatorname{argmin}}  \sum_{i=1}^K |C_i| \operatorname{Var} f(C_i),
   \end{aligned}
\end{equation}
where $\E[f(C_i)]$ is the mean embedding of points in cluster $C_i$.
Rather than solving the overall DML problem on $\mathbf{X}$, we now need to solve individual DML problems on the more homogeneous $C_i$. This reduced variance of the $C_i$ directly affects the sampling of negative and positive pairs from $C_i$ for DML: samples with different class labels (negatives) become harder, since the variance of $C_i$ and with that pairwise distances are smaller than for the entire dataset $\mathbf{X}$. Moreover, positives (samples with same class label) will be closer and thus easier.
Both effects are beneficial for training and improve model generalization \cite{xuan2020easy_positive}.
When the intra-class variance is high (e.g., classes with several visual modes, see Fig.~\ref{fig:cub_classes_splitted}), contracting all positive samples together will lead to mode collapse and the model will fail to distinguish visually dissimilar images within the same class, hindering the generalization to novel categories.
Thus, it is beneficial to push not all the positives together, but only easier ones \cite{xuan2020easy_positive} which would likely constitute the same mode. In Fig.~\ref{fig:cub_classes_splitted} we show different modes in training classes discovered by our division procedure.

Next, to achieve a smooth transition from an easy towards a more complex setup we
need to progressively increase the number of clusters and the corresponding number of subspaces every few epochs during training. We will introduce our division procedure by induction.
Given a current data partitioning into $K$ clusters $\mathbf{C}^{(t)}=\{C^{(t)}_1, \dots, C^{(t)}_K\}$,  where $t$ is the current division depth,
we need to divide it further in smaller subsets. The straightforward approach is to perform a hierarchical divisive clustering, e.g. Bisecting K-means \cite{bisecting_kmeans}, which divides every cluster $C^{(t)}_i$ in two subclusters $C^{(t+1)}_{2i-1}$ and $C^{(t+1)}_{2i}$.
However, such an approach leads to accumulating errors as we go deeper into the hierarchy, since earlier data splits are fixed and do not change during training.
To resolve this issue, before division, we update the current partition by reclustering  the data into $K$ clusters $\mathbf{C}^{\prime(t)}=\{C^{\prime(t)}_1, \dots, C^{\prime(t)}_K\}$ in the current embedding space, and only after that we divide every cluster into two subclusters with Bisecting K-means. This allows our method to benefit from constantly improving embedding representations during training.
However, to correctly re-assign embedding subspaces to the updated clusters (see Sect.~\ref{sec:emb_split}) we need to know the correspondences between the old clusters $C^{(t)}$ and the updated clusters $C^{\prime(t)}$. We compute these correspondences by solving a linear assignment problem
\begin{equation}\label{eq:linear_assignment}
    A = \underset{A}{\operatorname{argmin}} \sum_i \sum_j \operatorname{IoU}_{i,j} A_{i,j},
\end{equation}
where $\operatorname{IoU}_{i,j}$ is the similarity between an old cluster $C^{(t)}_i$ and an updated cluster $C^{\prime(t)}_j$
measured as intersection over union of the elements belonging to them, $A$ is a binary matrix with $A_{i,j} = 1$ iff $C^{(t)}_i$ is assigned to $C^{\prime(t)}_j$.
%





\begin{figure*}[t]
\begin{center}
 \includegraphics[width=0.95\linewidth]{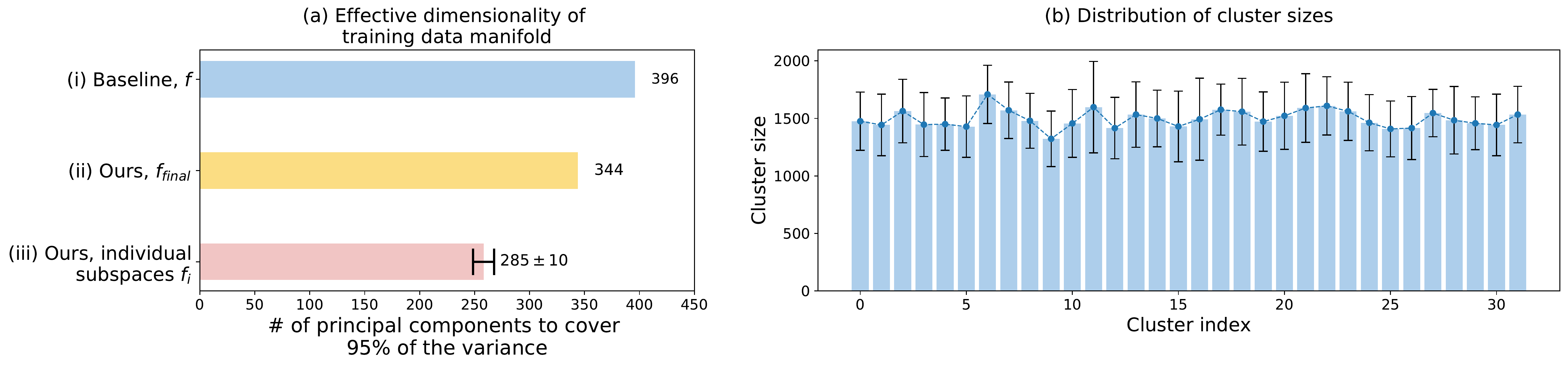}

\end{center}
   \caption{\textbf{(a)} Effective manifold dimensionality (ED) of the SOP training data in:
(i) the embedding space of the baseline with margin loss and $d=512$,
(ii) the final embedding space of our model.
Our model uses the embedding space more effectively and is able to learn lower dimensional embedding manifold compared to the baseline.
The ED decreases even more if we consider the individual clusters only after projecting them into their respective embedding subspaces (see (iii) for mean dimensionality over $32$ clusters).
\textbf{(b)} Distribution of the sizes of clusters obtained by our division approach during training on SOP. Error bars show mean and standard deviation of the sizes computed over different epochs when the clusters are updated.}

\label{fig:pca_components}
\end{figure*}

\subsubsection{Dividing the embedding space into subspaces}\label{sec:emb_split}
After we split the data into smaller clusters we assign every cluster to a subspace of the embedding space.
The naive approach would be to learn a new embedding layer from scratch for
each subset of the data, but this would imply an undesirable increase of the total embedding size and would prohibit sharing of information between the different embeddings.
Instead, we propose to split the original embedding space into subspaces, each selecting a subset of the original dimensions and masking out the rest.
%
This approach will enable sharing the representations between subspaces and will not blow up the overall embedding size.

Let us explain the masking in details.
At current division step $t$, for every cluster $C^{(t)}_i$  we learn a soft mask $s^{(t)}_i\in \mathbb{R}_+^d$ which induces a subspace by re-weighting individual embedding dimensions:
\begin{equation}\label{eq:emb_subspace}
    f^{(t)}_{i}(x) = f(x) \odot s^{(t)}_i,
\end{equation}
where $\odot$ stands for the element-wise product. \edited{Every mask $s^{(t)}_i$ is a learnable parameter.}  We use ReLU operator to truncate negative values in the masks.
Thus, $f^{(t)}_{i}$ is a mapping from the image space to the \emph{embedding subspace}. 
When we split a parent cluster $C^{(t-1)}_i$ into subclusters $C^{(t)}_{2i-1}, C^{(t)}_{2i}$, we create new masks $s^{(t)}_{2i-1}, s^{(t)}_{2i}$ for them which  are initialized using the mask of the parent, i.e.,
$s^{(t)}_{2i-1} := s^{(t-1)}_{i},  s^{(t)}_{2i} := s^{(t-1)}_{i}$.
Every subspace $f^{(t)}_{i}$ is then trained using non-overlapping data, thus learning different subspaces specifically tailored for their data.
Furthermore, to encourage stronger independence between learned subspaces we add an extra loss
\begin{equation}
    \mathcal{L}_{s} = \sum_{i\ne j} \frac{s^{(t)}_i \cdot s^{(t)}_j}{||s^{(t)}_i||_2 ||s^{(t)}_j||_2},
\end{equation}
which penalizes high correlation between subspaces.

\subsection{Training}\label{sec:emb_learning}

At current  division step $t$ with data partition $\mathbf{X} = \bigcup_{k=1}^K C^{(t)}_k$,  we optimize the embedding mapping $f$ by learning $K$ different embedding subspaces corresponding to the clusters.
All subspaces share the same underlying feature representation and therefore benefit from their mutual optimization.
We train the subspaces jointly in an alternating manner by randomly choosing a cluster $C^{(t)}_k,\, k \in \{1, \dots, K\}$ and drawing a random mini-batch $B_k$ from it. Then we need to update the $f^{(t)}_{k}$ to minimize the loss
\begin{align}\label{eq:learner_loss}
\begin{split}
    &\mathcal{L} = \sum_{k}\; \sum_{  (a,p,n) \sim B_k  }
    \left[ D_{a, p|s^{(t)}_k}^2 - D_{a, n|s^{(t)}_k}^2  + \alpha \right] +  \lambda \mathcal{L}_{s}, \\
\end{split}
\end{align}
\begin{equation}
    D_{i,j|s^{(t)}_k} = ||f^{(t)}_{k}(x_i) - f^{(t)}_{k}(x_j)||_2,
\end{equation}
where $(a, p, n) \in B_k \subset C^{(t)}_k$ is the triplet sampled from the current mini-batch $B_k$, and $D_{i,j|s^{(t)}_k}$ is the distance calculated in the subspace induced by the mask $s^{(t)}_k$.
The parameter $\lambda$ weights the contribution of the mask orthogonality loss.
We normalize $f^{(t)}_{k}(x_i)$ to unit length for training stability \cite{facenet}.

After each $E$ training epochs we perform the division again by splitting every cluster $C_k^{(t)}$ in $C_{2k}^{(t+1)}$, $C_{2k+1}^{(t+1)}$ and splitting the corresponding embedding subspace induced by mask $s_k^{(t)}$ by creating two new masks $s_{2k}^{(t+1)}$ and $s_{2k+1}^{(t+1)}$ (see Sec.\ref{sec:divide}).
Every division allows to achieve a deeper level of granularity and resolve more difficult relationships between images as we reduce the variance of the visual factors and focus on individual details. Once the desired number of sub-problems $K_{max}$ is reached, we continue training the model until convergence while also updating the data partitioning every $E$ epochs to account for the evolving embedding space. The full pipeline of our approach is visualized in Fig.~\ref{fig:pipeline}.

\subsection{Conquering}\label{sec:final_embedding}
For the final evaluation on the novel test categories we need a single measure of the similarity between images. Thus, we aim at combining the learned embedding subspaces together to produce a single mapping into the final embedding space, where the distances will be used to measure the similarity between test images. To produce a single mapping we sum the masks of the individual subspaces, which we then use to re-weight the dimensions of the entire embedding space:
\begin{equation}
    f_{final}(x) = f(x) \odot \sum_{i=1}^{K_{max}} s^{(t)}_i.
\end{equation}
Next, we fine-tune the final mapping to allow the subspaces to adjust to each other after summing their masks.
During this step, we fine-tune the network using the final embedding space for computing all the distances, but we still sample mini-batches from the clusters and not from the entire dataset to keep the hardness of the triplets on the same level.

\begin{table*}
    \centering

    \caption{Evaluation of different data division strategies on Stanford Online Products \cite{sop} ($K_{max}=32$), CUB200-2011 \cite{cub200_2011}, and CARS196 \cite{cars196} ($K_{max}=4$). Margin loss \cite{margin} and fixed orthogonal masks ("Ours (fix)") were used by default. \textbf{Progressive + Random data splits:} randomly and evenly split the data into clusters, while the data inside a cluster can only be further subdivided into child clusters.
    \textbf{Progressive + GT labels grouping:} manually group ground truth classes using the ground truth super-category annotations.
    \textbf{Not progressive:} divide the data directly into the maximal number of clusters $K_{max}$ w/o intermediate steps.
    }
    \label{tab:clustering_evaluation}

    \scalebox{0.7}{
    \begin{tabular}{l|ccccc||cccccc||ccccc}
    \toprule    \multicolumn{1}{l}{Datasets $\rightarrow$} &  \multicolumn{5}{c}{SOP} & \multicolumn{6}{c}{CUB200-2011}  & \multicolumn{4}{c}{CARS} \\    \midrule
    Scores    & R@1 & R@10 & R@100 & NMI & mARP & R@1 & R@2 & R@4 & R@8 & NMI & mARP & R@1 & R@2 & R@4 & NMI & mARP\\
    \midrule
    Margin loss & 77.11 & 89.33 & \textbf{95.57} & \textbf{89.65} & 76.98 & 66.93 & 77.43 & 85.15 & 90.92 & 69.12 & 54.07 & 85.12 & 91.07 & 94.87 & 69.31 & 63.45 \\

    \midrule

    Ours (fix) Progressive + Random data splits & 71.99 & 85.87 & 93.72 & 88.21 & 72.16 & 66.96 & 77.04 & 85.03 & 90.68 & 69.89 & 54.30 & 85.02 & 91.15 & 94.60 & 69.84 & 63.33\\

    Ours (fix) Progressive + GT labels grouping & 75.99 & 87.81 & 93.79 & 89.06 & 75.95 & 65.02 & 76.08 & 83.54 & 90.02 & 67.65 & 53.14 & 86.00 & 91.54 & 94.76 & 68.36 & 64.12\\


    \midrule
    Ours (fix) Not Progressive ~\cite{sanakoyeu_dcesml} & 78.67 & 89.53 & 94.77 & 89.48 & 78.56 & 67.18 & 77.23 & 85.04 & 90.80  & 69.60 & 55.26 & 86.08 & 91.23 & 94.48 & 69.37 & \textbf{64.58} \\
    \textbf{Ours (fix)} &  \textbf{79.54} & \textbf{90.24} & 95.13 & 89.58 & \textbf{79.35} & \textbf{68.43} & \textbf{78.68} & \textbf{85.85} & \textbf{91.32} & 69.74 & \textbf{55.47} & \textbf{86.46} & \textbf{91.66} & \textbf{94.97} & \underline{68.79 } & \underline{64.55}\\
    \bottomrule
    \end{tabular}
    }
\end{table*}

\begin{table*}
    \centering

    \caption{Evaluation of different subspace learning methods on Stanford Online Products, CUB200-2011, CARS196, and In-shop Clothes. \textbf{Ours (fix):} our method with fixed orthogonal masks. \textbf{Ours (learn):} our method with learnable masks.}
    \label{tab:subspace_masks_evaluation}

    \scalebox{0.7}{
    \begin{tabular}{l|ccccc||cccccc||ccccc||cccc}
    \toprule    \multicolumn{1}{l}{Datasets $\rightarrow$} &  \multicolumn{5}{c}{SOP} & \multicolumn{6}{c}{CUB200-2011}  & \multicolumn{5}{c}{CARS} & \multicolumn{4}{c}{In-shop} \\    \midrule
    Scores    & R@1 & R@10 & R@100 & NMI & mARP & R@1 & R@2 & R@4 & R@8 & NMI & mARP & R@1 & R@2 & R@4 & NMI & mARP &  R@1 & R@10 & NMI & mARP \\
    \midrule
    Margin loss & 77.11 & 89.33 & 95.57 & 89.65 & 76.98 & 66.93 & 77.43 & 85.15 & 90.92 & 69.12 & 54.07 & 85.12 & 91.07 & 94.87 & 69.31 & 63.45 & 87.87 & 96.26 & 89.16 & 86.14\\

    \midrule
    Ours (fix) No subspaces & 79.23 & 90.04 & 95.04 & 89.61 & 79.05 & 67.42 & 77.41 & 85.13 & 90.72 & 68.64 & 54.47 & 83.90 & 89.90 & 93.89 & 66.36 & 61.69 & 89.17 & 96.60 & 88.94 & 87.16\\

    \textbf{Ours (fix)} & 79.54 & 90.24 & 95.13 & 89.58 & 79.35 &  \textbf{68.43} & \textbf{78.68} & 85.85 & 91.32 & \textbf{69.74} & \textbf{55.47} & 86.46 & 91.66 & 94.97 & 68.79 & 64.55 & 90.15 & \textbf{97.62} & \textbf{89.96} & 88.15\\

    \textbf{Ours (learn)} & \textbf{79.77} & \textbf{90.39} & \textbf{95.20} & \textbf{89.67} & \textbf{79.56} & 68.16 & 78.14 & \textbf{85.97} & \textbf{91.64} & 69.49 & 55.35 & \textbf{87.76} & \textbf{92.52} & \textbf{95.35} & \textbf{70.67} & \textbf{65.97} & \textbf{90.40} & 97.52 & 89.91 & \textbf{88.47}\\
    \bottomrule
    \end{tabular}
    }
\end{table*}

\section{Experiments}\label{sec:experiments}
In this section, we evaluate our approach on five widely used benchmark datasets for fine-grained image retrieval and object re-identification. We quantitatively compare to the state-of-the-art methods and also provide qualitative results of our approach. To verify the significance of different parts of our model and the design choices we conduct ablation studies in Sec.~\ref{sec:abl_study}.

\subsection{Datasets}
For evaluation we use five standard benchmark datasets commonly used in deep metric learning.
Two smaller datasets: CARS196 \cite{cars196}, CUB200-2011 \cite{cub200_2011}; and three large-scale datasets: Stanford Online Products \cite{sop}, In-shop Clothes \cite{deepfashion}, and PKU VehicleID \cite{vehicleid}.
For the retrieval and re-identification tasks we report Recall@k \cite{recallk}, and for measuring the clustering performance we compute the normalized mutual information score (NMI) \cite{schutze2008introduction}.
As an alternative to NMI we also report mean Average R-precision (mARP) \cite{musgrave2020}.
For a detailed description of the datasets and evaluation metrics see supplementary material.

\subsection{Implementation Details}
We incarnate our approach using ResNet-50 \cite{resnet} architecture closely following the implementation of Wu et al.~\cite{margin}. We use dropout \cite{dropout} with probability $p=0.01$ before the embedding layer and set the embedding dimensionality to $d=512$ by default. The network is pretrained on Imagenet \cite{imagenet}, and the embedding layer is initialized using the ``Kaiming uniform'' initializer \cite{he2015initializers} which is the default initializer in Pytorch~\cite{pytorch}.

We set the maximal number of sub-problems $K_{max}=4$ for CUB200-2011 and CARS196 due to their small size.
For In-Shop Clothes we use $K_{max}=8$, for VID dataset $K_{max}=16$, and  for SOP $K_{max}=32$.
We divide (and recluster) the data every $10$ epochs for CUB200-2011 and CARS196 datasets, and every $2$ epochs for all the others.
In Sec.~\ref{sec:ablate_mod_epoch_and_k} we analyze the choice of $K_{max}$ and the number of epochs between division steps (denoted as $E$).
We resize the images to $256 \time 256$ and perform random cropping of size $224 \times 224$ and horizontal flips during training. We also experimented with some other augmentation strategies in Sec.~\ref{sec:augmentations}. A single central crop without augmentations is always used at test time.

In all experiments we use Adam \cite{adam} optimizer with the batch size of $112$ for CUB200-2011 \cite{cub200_2011} and CARS196 \cite{cars196}, and batch size of $80$ for the other datasets.
\wasanswered{For small datasets, namely CUB200-2011 and CARS196, we sample batches uniformly, and for all others we sample mini-batches following the procedure defined in \cite{facenet,margin} with $m=2$ images per class per mini-batch.}
We train all our models with a fixed learning rate of $10^{-5}$ without decay. The learning rate of the masks is scaled up $100$ times.
After 80 epochs for SOP and VID datasets we start fine-tuning the full embedding space. We observed that the fine-tuning does not bring an extra gain for smaller datasets like CUB200-2011, CARS196 and In-Shop Clothes.
We used cross-validation to tune the hyperparameters and report the detailed hyperparameter settings in the supplementary material.

\subsection{Ablation study}\label{sec:abl_study}
In this subsection we perform ablation experiments to evaluate different components of our approach.
\edited{Additionally, we experimented with fixed orthogonal masks. In this case every mask $s_i$ is initialized by setting only $d/K$ elements with indices $\{\frac{(i-1)d}{K}, \dots, \frac{i d}{K} - 1\}$ to be one and everywhere else zero and not updated.}
We fix the subspace masks for all the ablation experiments if not stated differently.


\subsubsection{Division strategy}

First, to validate that the proposed progressive division strategy is important, we substitute it with a division of the data directly into the maximal number of clusters $K_{max}$ and learning $K_{max}$ different subspaces on them (``Not progressive''). However, we still recluster the data every $E$ epochs in this case. In Tab.~\ref{tab:clustering_evaluation} we can see that this results in a big drop in recall and mARP, since the model tries to solve substantially harder sub-problems without learning how to solve simpler intermediate problems.
Fig.~\ref{fig:plot_loss_nmi} (b) shows that the metric learning loss is higher for our method compared to the baseline and it even slightly grows after the $6$-th epoch  because the number of sub-problems increases during the first $12$ epochs from $1$ to $32$. Hence, starting with $32$ sub-problems from the onset does not allow the model to be trained in the easier regime at the beginning.

Second, we study the importance of using the learned embedding space for clustering the data.
Instead of clustering we partition the data every $E$ epochs by random bisecting splits which do not utilize the embedding space (see Tab.~\ref{tab:clustering_evaluation}, ``Progressive + Random data splits''). This decreased the recall by $7.5\%$ for SOP and by $1.5\%$ for CUB200-2001 and CARS196 showing that it is important to split the data according to the current embedding space configuration.
We also manually grouped ground truth classes into clusters using the ground truth super-category annotations provided with SOP \cite{sop}, CUB200-2001 \cite{cub200_2011} and CARS196 \cite{cars196} datasets (see Tab.~\ref{tab:clustering_evaluation}, ``Progressive + GT labels grouping''). This works better than random dataset partitioning, but is still much worse than our approach, since the predefined data partitioning does not consider what the embedding space had already learned and does not allow adaptive zooming in on the most challenging parts of the data. For CUB200-2001 such ground truth grouping is even worse than random. We speculate that the reason for this is that CUB200-2001 is very fine-grained dataset, and it is very hard to group the super-categories manually. In Fig.~\ref{fig:plot_loss_nmi} (c) one can see that the clusters produced by the proposed approach change during training and do not necessarily correspond to the ground truth (NMI with GT classes is around $0.4$)
.
The blue curve in Fig.~\ref{fig:plot_loss_nmi} (c) demonstrates NMI between cluster memberships of the samples before and after each division step. As we can see, NMI remains in the range of $[0.35, 0.6]$, which means that around $60\%-75\%$ of the points retain the same cluster label after solving the linear assignment problem defined in Eq.~\ref{eq:linear_assignment}. NMI value close to $1.0$ means that the clusters do not change, which would be equivalent to using fixed data partitioning (see Tab.~\ref{tab:clustering_evaluation}, ``Progressive + GT labels grouping") resulting in a decrease of R@1 by $3.5\%$. On the other hand, drastic changes of clusters would result in even higher loss of performance (e.g., in case of random data splits we observe a $7.5\%$ drop in R@1, see Tab.~\ref{tab:clustering_evaluation} ``Progressive + Random data splits"). Therefore, we can conclude that in the proposed method cluster memberships of samples change smoothly enough for the method to outperform both fixed clustering and random data partition.
Fig.~\ref{fig:pca_components} (b) shows that we get clusters of approximately equal sizes.

\subsubsection{Subspace learning}

\begin{table}[t]
    \centering
    \caption{Evaluation of the models trained with different division frequency on SOP dataset (fixed masks).}
    \scalebox{0.85}{
    \begin{tabular}{lccccc}
        \toprule
        \textbf{Epochs btw. divisions, $E$} & 1 & 2 & 5 & 10 & 20\\
        \midrule
        \textbf{R@1} & 79.01 & \textbf{79.54} & 78.68 & 78.24 &  77.22 \\
        \textbf{NMI} & 89.55 & \textbf{89.58} & 89.48 & 89.45 & 89.24 \\
        \textbf{mARP} & 78.82 & \textbf{79.35} & 78.53 & 78.05 & 77.09 \\
        \bottomrule
    \end{tabular}
    }
    \label{tab:reclustering_freq}
\end{table}

\begin{table}[t]
\caption{Evaluation of the models trained with different maximal number of sub-problems $K_{max}$ on SOP dataset (fixed masks).}
    \setlength\tabcolsep{2.0pt}
    \centering
    \scalebox{0.9}{
        \begin{tabular}{lccccccc}
        \toprule
        \textbf{Sub-problems $K_{max}$} & 1 & 2 & 4 & 8 & 16 & 32 & 64\\
        \midrule
        \textbf{R@1} & 77.11 & 78.25 & 78.71 & 79.19 & 79.28 & \textbf{79.54} & 79.51\\
        \textbf{NMI} & 89.65 & 89.88 & \textbf{89.89} & 89.79 & 89.70 & 89.58 & 89.42 \\
        \textbf{mARP} & 76.98 & 78.10 & 78.53 & 78.96 & 79.14 & \textbf{79.35} & 79.31 \\
        \bottomrule
        \end{tabular}
    }
    \label{tab:k_max}
\end{table}

\begin{figure}[t]
\begin{center}
 \includegraphics[width=0.70\linewidth]{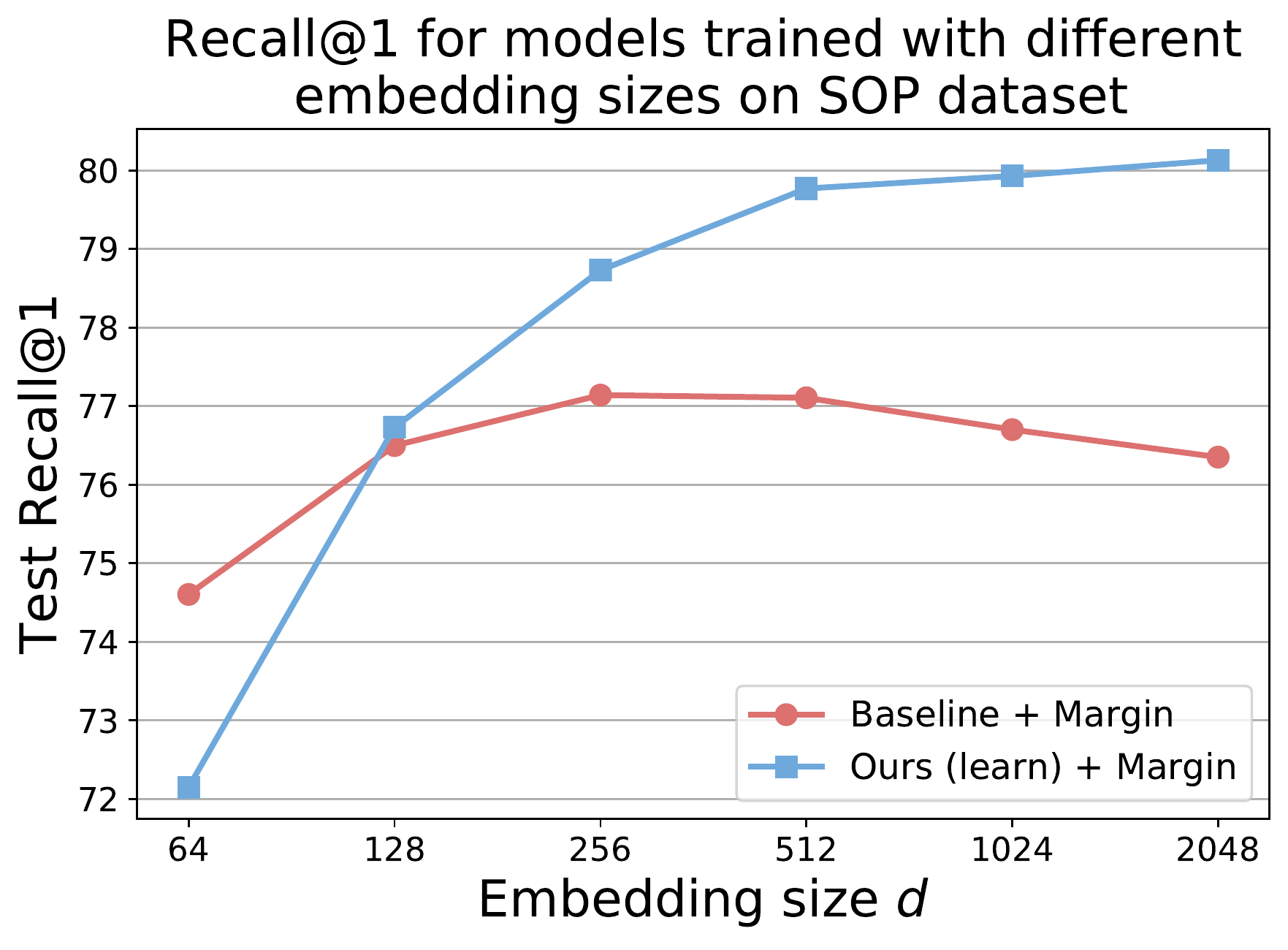}

\end{center}
   \caption{Recall@1 score for our approach and the baseline margin loss method \cite{margin} trained using different embedding sizes $d$ on SOP dataset. In contrast to the baseline method, our approach enables more efficient utilization of larger embeddings and does not overfit even for a very large embedding size.
   }
\label{fig:ablations_emb_size}
\end{figure}

\begin{figure}[t]
\begin{center}
 \includegraphics[width=0.75\linewidth]{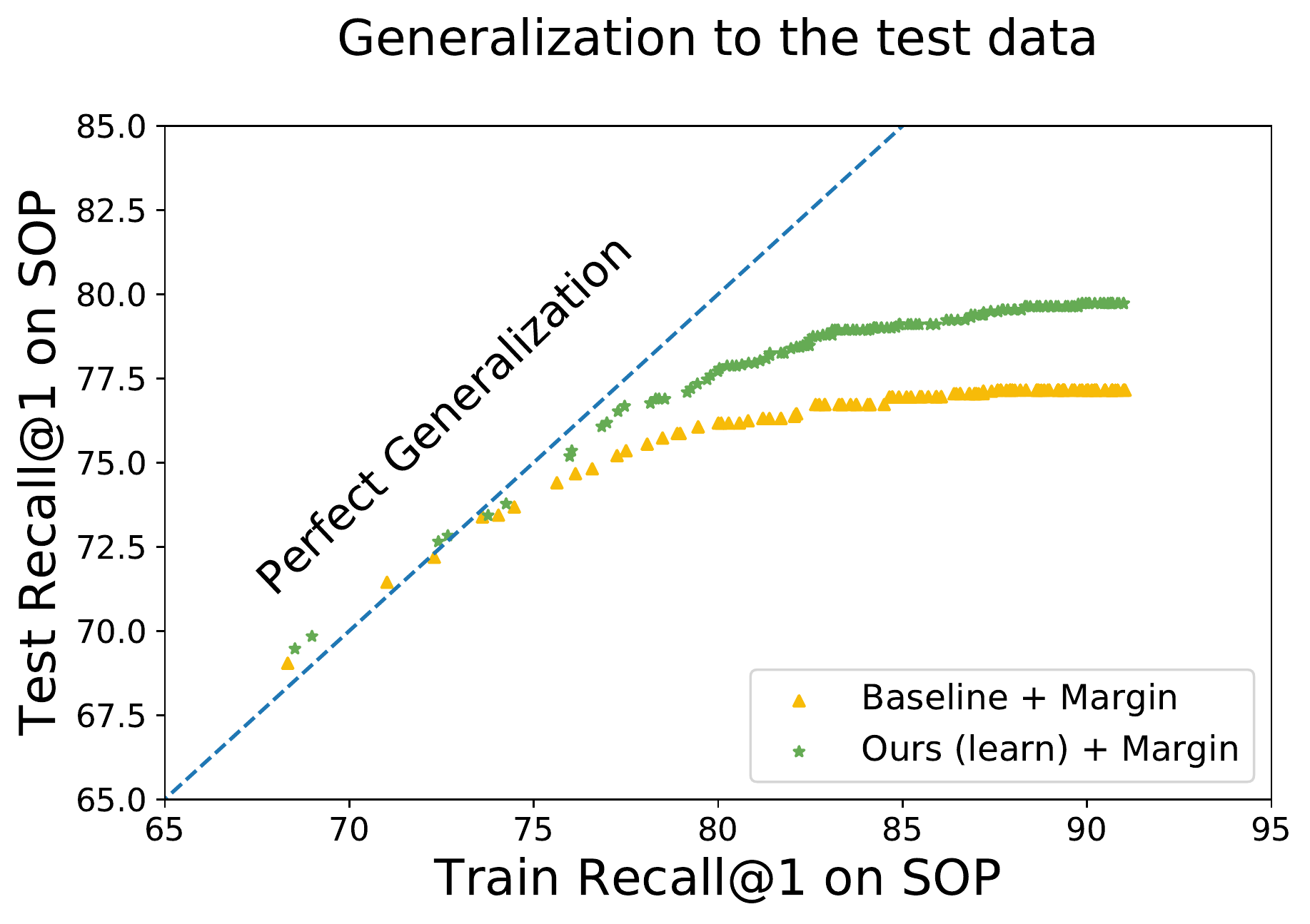}

\end{center}
   \caption{We visualize Train Recall@1 vs Test Recall@1 on SOP dataset for every training epoch. The plot shows that the proposed model ($K_{max}=32$) better generalizes on previously unseen test classes and displays less overfitting to train classes compared to the baseline ($K_{max}=1$). We omitted the first $2$ epochs for compactness of the figure.
   }
\label{fig:train_vs_test_recall}
\end{figure}

In Tab.~\ref{tab:subspace_masks_evaluation} we evaluate different variants for subspace learning.
We compare the use of learnable masks $s_k$
and fixed masks which are initialized to be orthogonal.
Learnable masks outperform fixed ones everywhere except for CUB200-2011, where they perform approximately on par. We explain such behaviour by a slight overfitting of the learnable masks when the dataset is too small like in the case of CUB, which contains only $5864$ training images. Visualizations of the learned masks are shown in the supplementary material.
To show that learning different subspaces for different data clusters is important we disable this component and train the entire embedding space regardless of the chosen cluster. In Tab.~\ref{tab:subspace_masks_evaluation} (``No subspaces'') we can observe that training without separate subspaces leads to a substantial performance drop.

In Fig.~\ref{fig:cmp_learners_vs_baseline_cars} we show the retrieved nearest neighbors in different subspaces and in the final combined space. We noticed that subspace  $\#3$ makes similar mistakes as the baseline model (confuses a GMC van with a Ford Mustang), however, different subspaces are able to correct each other which leads to more accurate results when we combine them together.

We estimated the effective dimensionality (ED) of the data manifold in the learned embedding space as the number of principal components required to cover the $95\%$ of the variance. In other words, ED defines the minimal dimensionality of such a linear manifold that the projection on it encodes almost all ($95\%$) of the data variance.
In Fig.~\ref{fig:pca_components} we show that the ED of the data manifold
in the embedding space of our model is significantly lower than the ED of the data for the baseline model ($K_{max}=1, d=512$).
Lower ED and higher recall implies that our model provides higher compression of the data and encodes more structure in the embedding space than the baseline.
Moreover from Fig.~\ref{fig:pca_components} (iii) we see that every data cluster after projecting in the corresponding subspace has even lower ED than ED for the full dataset which shows that the learned subspaces effectively encode and compress the data samples from the corresponding clusters.

\begin{figure*}[t]
\begin{center}
\includegraphics[width=0.96\linewidth]{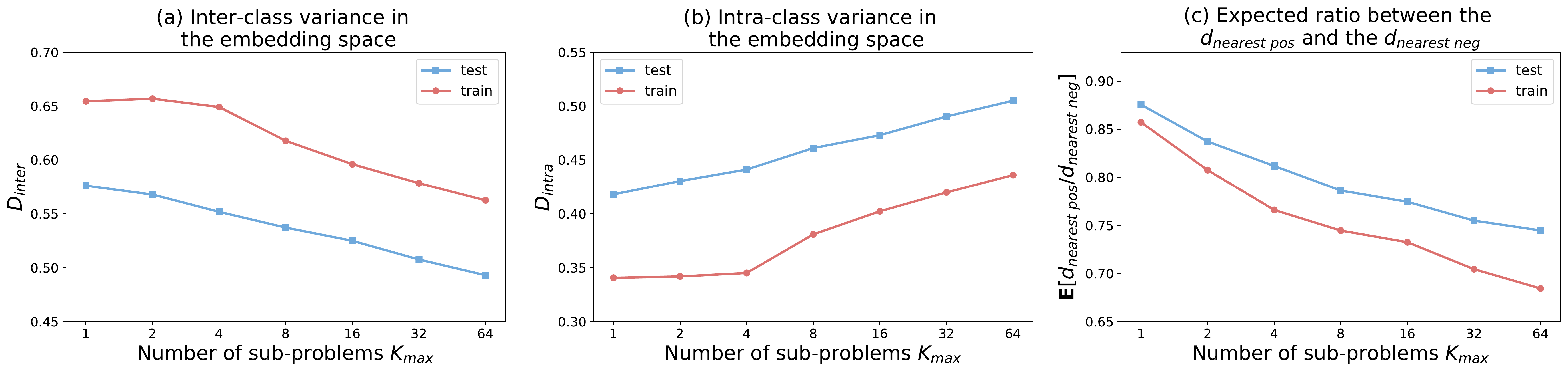}

\end{center}
   \caption{\textbf{(a) and (b)} Inter- and intra-class variance for our models with different number of sub-problems $K_{max}$ trained with margin loss \cite{margin} on SOP dataset. $K_{max} = 1$ is the baseline.
   \textbf{(c)} Expected ratio between the distance to the nearest example from the same class and the distance to the nearest example from a different class. The lower this ratio, the more likely that the nearest neighbor has the same class label. Intra-class variance grows with $K_{max}$ because it becomes more likely for a class to be splitted into different modes by our division step. Hence, with higher $K_{max}$, classes become less compact, with more overlap in their periphery, because our method does not enforce all positives to contract together.
   }
\label{fig:plot_variance}
\end{figure*}

\begin{figure*}[t!]
\begin{center}
\includegraphics[width=0.99\linewidth]{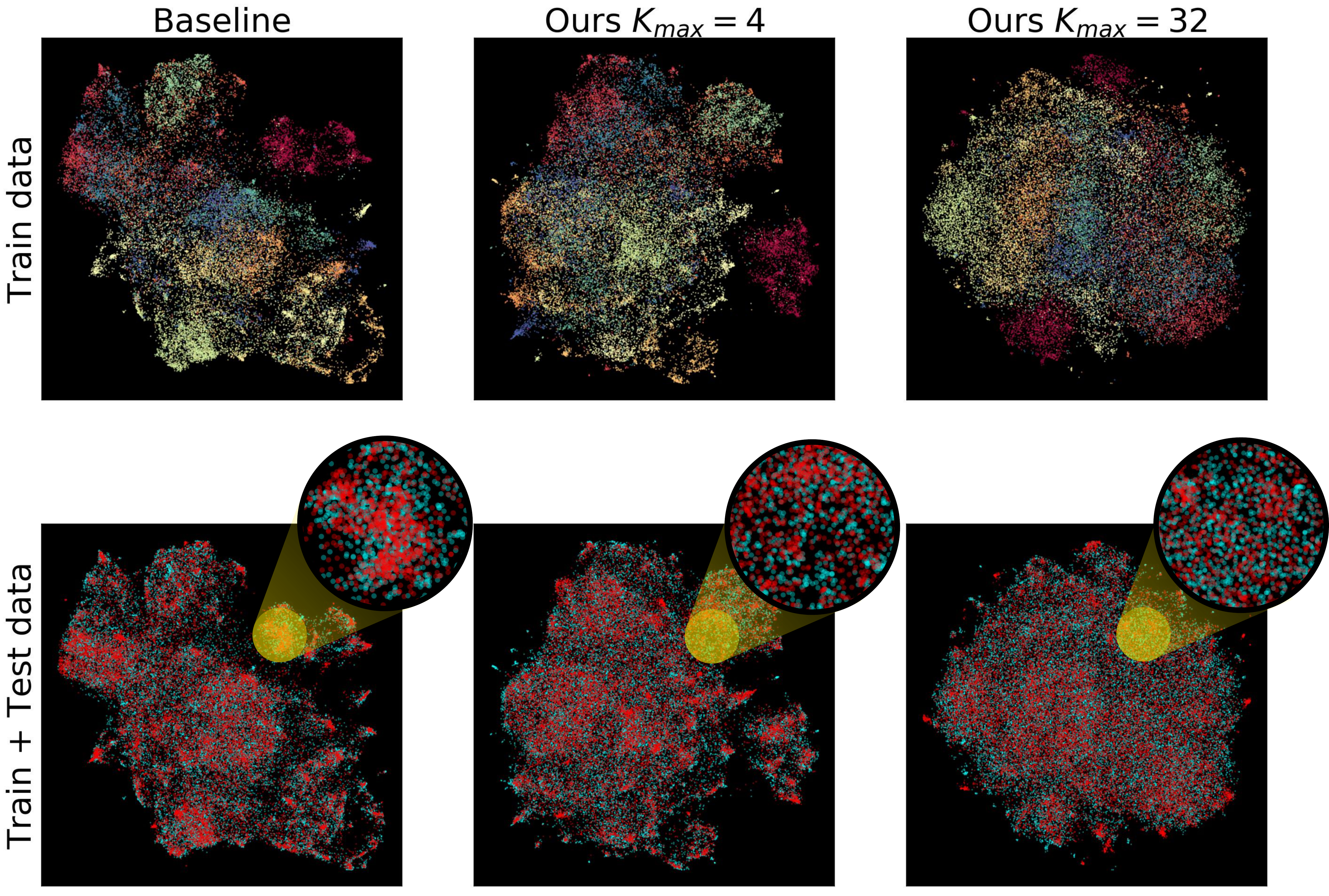} 
\end{center}

 \definecolor{umapblue}{RGB}{5,243,247}
 \definecolor{umapred}{RGB}{228,13,14}

  \caption{Low-dimensional projection of the embedding vectors produced by UMAP \cite{umap} for models trained on the SOP dataset with margin loss. We compare baseline ($K_{max}=1$) to our models with $K_{max}=4$ and $K_{max}=32$. \textbf{(1-st row)} we visualize embedding for training data, where color encodes class label ($11318$ classes); \textbf{(2-nd row)} embedding for \emph{train} (in \emph{blue}) and \emph{test} data (in \emph{red}) together. The embeddings produced by our model with $32$ clusters have fewer high-density hubs and are more uniformly distributed in the space while still maintaining the superior retrieval performance. We computed expected average distance between sample and its ten nearest neighbors in the embedding space ($ED_{10}$) for the baseline model on train ($0.918$) and on test ($0.953$). $ED_{10}$ increases for our model with $K_{max}=32$ to $1.044$ and $1.069$ correspondingly. While the $ED_{10}$ is becoming closer to a uniform distribution on the unit sphere ($1.290$), the distance to nearest positive decreases relative to the distance to nearest negative.
  This provides an extra evidence that our model  utilizes the embedding space in its entirety more effectively.
  Zoom-in for details.}
\label{fig:umap}
\end{figure*}

\begin{table*}[t]
    \centering
    \caption{Performance on different datasets using various image augmentation settings.}
    \scalebox{0.85}{
    \begin{tabular}{ll|ccc||ccc||ccc||ccc}
        \toprule
        & \multicolumn{1}{l}{Datasets $\rightarrow$} & \multicolumn{3}{c}{CUB200-2011} & \multicolumn{3}{c}{CARS196} & \multicolumn{3}{c}{In-shop} & \multicolumn{3}{c}{SOP} \\
        \midrule
        & Augmentations $\downarrow$ & R@1 & NMI & mARP & R@1 & NMI & mARP & R@1 & NMI & mARP & R@1 & NMI & mARP\\
        \toprule
        \multirow{4}{*}{\rotatebox{90}{Baseline}} & Standard  & 66.24 & 68.46 & 53.63  & 84.15 & 69.09 & 62.22  & 87.11  & 88.79 & 85.33 & \textbf{77.11} & \textbf{89.65}  & \textbf{76.98}  \\
        & Geom & 66.31 & 67.63 & 53.56 & 81.23 & 66.80 & 58.65 & \textbf{87.87} & \textbf{89.16} & \textbf{86.14} & 76.14 & 89.48 & 75.94\\
        & Geom+Color & \textbf{66.93} & \textbf{69.12}& \textbf{54.07} & 81.24 & 65.43 & 58.16 & 87.04 & 88.70 & 85.26 & 75.84 & 89.47 & 75.75\\
        & Color Heavy & 65.68 & 68.63 & 53.62 & \textbf{85.12} & \textbf{69.31} & \textbf{63.45} & 79.02 & 87.53 & 78.95 & 72.17 & 88.62 & 72.42\\
        \midrule

        \multirow{4}{*}{\rotatebox{90}{Ours (fix)}} & Standard  & 67.45 & 69.71 & 54.92 & 85.77 & 69.93 & 64.35 & 88.67 & 89.25 & 86.89  & \textbf{79.54} & \textbf{89.58} & \textbf{79.35} \\

        & Geom & 67.35 & 69.13 & 54.04 & 84.49 & 68.75 & 62.51 & \textbf{90.15} & \textbf{89.96} & \textbf{88.15}  & 78.82 & 89.50 & 78.63\\

        & Geom+Color & \textbf{68.43} & \textbf{69.74}  & \textbf{55.47} & 84.68 & 67.64 & 62.61  & 88.35 & 89.31 & 86.63 & 78.48 & 89.41 & 78.34 \\

        & Color Heavy & 66.26 & 70.27 & 54.59 &  \textbf{86.46} & \textbf{68.79} & \textbf{64.55} & 82.22 & 87.99 & 81.56 & 76.02 & 88.77 & 76.03 \\
        \bottomrule
    \end{tabular}}
    \label{tab:augmentations}
\end{table*}

\subsubsection{Embedding size}

We also tested how our approach performs for various embedding sizes.
In Fig.~\ref{fig:ablations_emb_size} we show the Recall@1 score for our approach and the baseline with margin loss \cite{margin} for different embedding sizes on
the SOP dataset. Interestingly, the baseline cannot benefit from a larger embedding size due to severe overfitting. Existing works also report a decrease in performance when they train the embedding space with more than $256$ \cite{lifted_struct} or $512$ \cite{wang2019multisim_loss,roth2019mic,tuplet_margin_loss2019} dimensions. In contrast, our approach gains the performance when we increase the embedding space size due to the more efficient distribution of the learned information across embedding dimensions by simultaneous division of the data and the embedding space into sub-problems.

We observe that our approach performs worse with the embedding size of $64$ due to insufficient embedding capacity for learning $32$ independent subspaces. However, the performance still improves even for such a large embedding size as $2048$ (see Fig.~\ref{fig:ablations_emb_size}).

\subsubsection{Division frequency and the number of sub-problems}\label{sec:ablate_mod_epoch_and_k}

Tab.~\ref{tab:reclustering_freq} varies the number of training epochs $E$ between consecutive divisions. In Tab.~\ref{tab:k_max} we change the maximal number of sub-problems $K_{max}$. It can be seen that the best performance for SOP is achieved when we divide every $2$ epochs and the maximal number of sub-problems is $32$. Moreover, in Tab.~\ref{tab:k_max} one can see how R@1 and mARP continuously grow when we increase $K_{max}$ from $1$ to $32$.

In Fig.~\ref{fig:plot_variance} we see that increasing the number of clusters and subspaces leads to higher intra-class variance because it becomes more likely for a class to be splitted into different visual modes by our division step. Since different visual modes within a class are not enforced to contract together anymore, the classes become less compact and inter-class variance decreases. However, this promotes generalization and improves retrieval on the test set: The higher $K_{max}$, the more likely is it for a given query to retrieve the nearest neighbor with the same class label (see Fig.~\ref{fig:plot_variance} (c)).

Fig.~\ref{fig:umap} displays 2D projections of the learned image embeddings.
The baseline model overfits to the training classes (Fig.~\ref{fig:umap}, first column)
resulting in high-density hubs in the test embedding distribution.
However, for our approach (Fig.~\ref{fig:umap}, rightmost column), the embedded test data is more spread over the embedding space and not crammed in few locations, thus better matching the training distribution compared to the baseline. Average distance between embedded samples and their ten nearest neighbors ($ED_{10}$) increases from $0.953$ (baseline) to $1.069$ (ours with $K_{max} = 32$), approaching the $ED_{10}$ for uniformly distributed random vectors on a unit sphere ($1.290$). Importantly however, while the average distances between the embeddings increase, the distance to the nearest positive sample decreases relative to the distance to the nearest negative sample (See Fig.~\ref{fig:plot_variance} (c)).
This highlights that our approach can better utilize the embedding space and is subject to less overfitting to the training data, which is also shown in  Fig.~\ref{fig:train_vs_test_recall}.

\subsubsection{Image augmentations}\label{sec:augmentations}

We further experimented with different augmentation strategies: (Standard) the standard augmentation strategy with random cropping of size $224\times×224$ and horizontal flips; (Geom) the same as ``standard'' plus random rotations, shifts, blur and cutouts \cite{2017cutout} with probability $p=0.5$; (Geom+Color) the same as ``Geom" plus random changes of brightness and contrast with $p=0.5$; (Color Heavy) the same as ``standard" plus random color jitter with $p=0.8$ and conversion to the grayscale with $p=0.2$.

Different datasets require different augmentations. CARS196 benefit from heavy color augmentations because what matters in this dataset is the car model, not its color. In contrast, for In-shop Clothes the model is required to retrieve photos of exactly the same fashion item and the color is obviously a very important clue. Therefore, applying heavy color augmentations deteriorates the performance drastically. One can see that our approach is more robust to such harmful augmentations. The recall@1 score of the baseline dropped by $8.1\%$ while the recall@1 of our model decreased only by $6.4\%$ (cf. Tab.~\ref{tab:augmentations}, ``standard" and ``Color Heavy"). For CUB200-2001 geometrical and mild color augmentations are the most suitable because birds of the same class have different poses and can have slightly different shades. For large-scale datasets like SOP and PKU VehicleID, augmentations do not bring extra performance, potentially due to already high diversity of the images in the dataset. In the next sections, we report the performance our models and baselines trained with the best augmentation strategies for every dataset.

\begin{table*}[t]
 \rowcolors{2}{gray!8}{white}
   \centering
   \caption{Comparison of our approach with the state-of-the-art methods. We report Recall@1, NMI, and mARP on CUB200-2011, CARS196, and SOP datasets.
   $\dagger$ denotes our implementation;
   * denotes ensembling methods or methods which require non-standard network architecture modifications.
   }

    \label{tab:sota_results_cub_cars_sop}

   \scalebox{0.9}{ 
   \begin{tabular}{l|lc||ccc||ccc||ccc}

     \toprule

     \multicolumn{3}{l}{Datasets $\rightarrow$} &
     \multicolumn{3}{c}{CUB200-2011} &
     \multicolumn{3}{c}{CARS196} &
     \multicolumn{3}{c}{SOP} \\
     \midrule

     Approaches $\downarrow$ & Architecture & Dim & R@1 & NMI & mARP & R@1 & NMI & mARP & R@1 & NMI & mARP \\

    \midrule
    DAML (N-pairs) \cite{daml} & GoogleNet & 512 & 52.7 &  61.3 & - & 75.1 & 66.0 & - & 68.4 & 89.4 & -\\
    HDML (N-pairs) \cite{hdml} & GoogleNet & 512 & 53.7 &  62.6 & - & 79.1 & 69.7 & - & 68.7 & 89.3 & -\\
    Angular \cite{angular} & GoogleNet     & 512 & 54.7 &  61.1 & - & 71.4 & 63.2 & - & 70.9 & 88.6 & -\\
    HDC \cite{hdc} & Googlenet             & 384 & 53.6 &  -    & - & 73.7 & -    & - & 69.5 & - & -\\
    BIER \cite{bier_iccv} & GoogleNet      & 512 & 55.3 &  -    & - & 78.0 & -    & - & 72.7 & - & -\\
    HTL \cite{htl} & BNInception           & 512 & 57.1 &  -    & - & 81.4 & -    & - & 74.8 & - & -\\
    A-BIER \cite{a_bier} & GoogleNet       & 512 & 57.5	&  -    & - & 82.0 & -    & - & 74.2 & - & -\\
    Smart Mining~\cite{smart_mining,suh2019softmax_and_triplet} &  GoogleNet & 512 & 50.2 & - & - & 72.7 & - & - & - & - & -\\
    Stochastic Mining~\cite{suh2019softmax_and_triplet} & GoogleNet & 512 & 55.1 & - & - & 82.5 & - & - &72.1 & - & -  \\
    HTG \cite{hard_triplet_gen} & ResNet-18  & 512 & 59.5 & -    & - & 76.5 & -    & - & - & - & -\\
    Tuplet Margin~\cite{tuplet_margin_loss2019} & ResNet-50 & 512 & 62.5 & - & - & 86.3 & - & - & 78.0 & - & - \\
    NormSoftmax~\cite{zhai2018making_normsoftmax} &  ResNet-50 & 512 & 61.3 & - & - & 84.2 & - & - & 78.2 & - & - \\
    ICE~\cite{wang2019instance} & BNInception & 512 & 58.3 & - & - & 77.0 & - & - & 77.3 & - & -\\
    RankedList~\cite{rankedlist2019} & BNInception & 512 & 57.4 & 63.6 & - & 74.0 & 65.4 & - & 76.1 & 89.7 & -\\
    SoftTriple~\cite{qian2019SoftTriple} & BNInception & 512 & 65.4 & 69.3 & - & 84.5 & 70.1 & - & 78.3 & 92.0 & -\\
    MultiSim~\cite{wang2019multisim_loss} & BNInception & 512 & 65.7 & - & - & 84.1 & - & - & 78.2 & - & - \\
    EPSHN~\cite{xuan2020easy_positive} &  ResNet-50 & 512 & 64.9 & - & - & 82.7 & - & - & 78.3 & - & -\\
    FastAP~\cite{cakir2019fastAP} & ResNet-50 & 512 & - & - & - & - & - & - & 75.8 & - & -\\
    \midrule
    ABE-8\cite{kim2018_abe} & GoogleNet* & 512 & 60.6 & - & - & 85.2 & - & - & 76.3 & - & -\\
    EDMS~\cite{EDMS_2019_CVPR} & ResNet-18$\times 25$* & 3200 & 66.1 & 68.9 & - & 87.6 & 76.7 & - & 78.5 & 90.1 & -\\
    DREML \cite{dreml} & ResNet-18$\times 48$* & 576  & 63.9 & 67.8 & - & 86.0 & 76.4 & - & - & - & -\\
    \midrule
    Triplet$\dagger$  & ResNet-50 & 512 & 62.44 & 65.52 & 51.21 & 75.30 & 59.66 & 53.88 & 76.59 & 90.03 & 76.51\\
    \edited{Ours (fix) Not Progressive + Triplet$\dagger$} & ResNet-50 & 512 & 64.89 & 68.02 & 53.59 & 76.52 & 63.77 & 56.54 & 77.23 & 90.00 & 77.03 \\%
    \edited{\textbf{Ours (fix) + Triplet}$\dagger$} & ResNet-50 & 512 & 65.01 & 67.61 & 53.27 & 77.17 & 62.95 & 56.79 & 77.46 & 90.22 & 77.27 \\%
    \textbf{Ours (learn) + Triplet}$\dagger$ & ResNet-50 & 512 & 64.33 & 67.89 & 53.14 & 77.51 & 64.13 & 57.20 & 79.05 & 90.43 & 78.83 \\
    \midrule
    Soft Triplet$\dagger$  & ResNet-50 & 512 & 61.60 & 65.66 & 50.71 & 72.18 & 62.17 & 52.73 & 71.04 & 88.91 & 71.26 \\
    \edited{Ours (fix) Not Progressive + Soft Triplet$\dagger$} & ResNet-50 & 512 & 64.89 & 67.42 & 53.27 & 76.37 & 63.03 & 55.98 & 77.04 & 90.06 & 76.80 \\
    \edited{\textbf{Ours (fix) + Soft Triplet}$\dagger$} & ResNet-50 & 512 & 65.24 & 67.45 & 53.22 & 75.81 &  62.55 & 55.78 & 77.12 & 90.12 & 76.94 \\
    \textbf{Ours (learn) + Soft Triplet}$\dagger$ & ResNet-50 & 512 & 64.87 & 67.47 & 53.27 & 76.73 & 63.23 & 55.80 & 77.59 & 90.16 & 77.38\\
    \midrule
    N-pairs$\dagger$ & ResNet-50 & 512 & 62.22 & 65.69 & 50.77 & 77.11 & 64.52 & 56.36 & 78.15 & 89.96 & 77.94\\
    \edited{Ours (fix) Not Progressive + N-pairs$\dagger$} & ResNet-50 & 512 & 63.52 & 66.40 & 52.11 & 77.58 & 64.13 & 56.79 & 76.79 & 89.85 & 76.69 \\
    \edited{\textbf{Ours (fix) + N-pairs}$\dagger$} & ResNet-50 & 512 & 63.94 & 65.51 & 51.79 & 78.54 & 65.11 & 57.81 & 80.14 & 90.23 & 79.82\\
    \textbf{Ours (learn) + N-pairs}$\dagger$ & ResNet-50 & 512 & 63.83 & 65.74 & 51.89 & 79.41 & 65.63 & 58.37 & \textbf{80.32} & \textbf{90.22} & \textbf{80.02} \\
    \midrule

    Margin$\dagger$  & ResNet-50 & 512 & 66.93 & 69.12 & 54.07 & 85.12 & 69.31 & 63.45 & 77.11 & 89.65 & 76.98\\
    Ours (fix) Not Progressive + Margin$\dagger$~\cite{sanakoyeu_dcesml} & ResNet-50 & 512 &
    67.18 & 69.60 & 55.26 &
    86.08 & 69.37 & 64.58 &
    78.67 & 89.48 & 78.56 \\
    \textbf{Ours (fix) + Margin}$\dagger$ & ResNet-50 & 512 & \textbf{68.43} & \textbf{69.74} & \textbf{55.47} & 86.46 & 68.79 & 64.55 & 79.54 & 89.58 & 79.35 \\
    \textbf{Ours (learn) + Margin}$\dagger$ & ResNet-50 & 512 & 68.16 & 69.49 & 55.35 & \textbf{87.76} & \textbf{70.67} & \textbf{65.97} &  \textbf{79.77} & \textbf{89.67} & \textbf{79.56}\\

    \bottomrule

    \end{tabular}
    }

 \end{table*}

\subsubsection{Different loss functions}
Our approach is applicable independently of a particular deep metric learning method used to solve the sub-problems.
To demonstrate this, we experimented with four different loss functions: Margin loss \cite{margin}, triplet loss \cite{facenet}, soft triplet loss \cite{npairs,hermans2017in_defense_triplet_loss}, and N-pairs loss \cite{npairs}.
We set the margin parameter $\alpha$ to $0.2$ for triplet loss \cite{facenet}, and $\beta=1.2$ for margin loss as proposed in \cite{margin}.
In Tab.~\ref{tab:sota_results_cub_cars_sop}, our method shows a consistent improvement for all four tested loss functions. We obtain up to $5.3\%$ relative increase in R@1 compared to the baseline for CUB200-2011, up to $6.3\%$ for CARS196, and up to $9\%$ for SOP. We observe especially large performance improvement when our method is used with soft triplet loss \cite{hermans2017in_defense_triplet_loss}.

\subsubsection{Runtime complexity:}
We cluster the data every $E$ epochs. To do this we use the linear-time K-means implementation from the FAISS \cite{faiss} library which has an average complexity of $O(K n i)$, where $n$ is the number of samples, $K$ is is the number of clusters, and $i$ is the number of iterations. Clustering of $100000$ $512$-dimensional vectors with K-means takes approximately five seconds, which is a neglectable amount compared to the time required for the full forward and backward pass of the entire dataset through the network. To compute the embeddings used for clustering, only one forward pass on the full dataset is needed, which amounts to an overhead of $\approx 5\%$ in case of $E=10$. \edited{The linear assignment problem in Eq.~\ref{eq:linear_assignment} is solved using Jonker-Volgenant algorithm \cite{jonker1987shortest} which has a complexity of $O(K^3)$ and takes less than a second to run.}


\subsection{Comparison with state-of-the-art}

In this section, we compare our approach to state-of-the-art methods.
Stochastic Mining \cite{suh2019softmax_and_triplet} uses large image size ($336\times336$), second order pooling \cite{compact_bilinear_pool2016} and an extra classification loss which greatly increase the model performance. Our method can further benefit from these extra modifications as well. However, for fair comparison, we compare only to the version of \cite{suh2019softmax_and_triplet} which used a standard image size of $224\times224$, average pooling and which did not back-propagate from the classification loss to the network backbone. We do not compare with \cite{kim2020proxy_anchor} because, in contrast to our implementation, they used a non-standard modification of ResNet-50 with a sum of global average pooling and global max pooling before the final layer, which brings a significant boost in performance making it hard to compare with other existing methods.
For the FastAP \cite{cakir2019fastAP} method we report the performance obtained by training with the batch size $96$ to have comparable batch size with our models which did not use large-batch training heuristics.

\begin{table*}
    \rowcolors{2}{gray!8}{white}
    \centering
    \caption{Recall@k on the small, medium, and large test sets of PKU VehicleID \cite{vehicleid} dataset. $\dagger$~denotes our implementation with the embedding size of $512$; *~denotes ensembling methods or methods which require non-standard network architecture modifications.}
    \label{tab:vehicleid}

    \scalebox{0.9}{
    \begin{tabular}{lcccc||cccc||cccc}
    \toprule
    Split Size & \multicolumn{4}{c}{Small}  & \multicolumn{4}{c}{Medium} & \multicolumn{4}{c}{Large} \\
    \midrule
    R@k    & 1 & 5 & NMI & mARP & 1 & 5 & NMI & mARP  & 1 & 5 & NMI & mARP \\
    \midrule
    GS-TRS loss \cite{em2017incorporating} & 75.0 & 83.0 & -  & -  & 74.1 & 82.6 & -  & -  & 73.2 & 81.9 & -  & - \\
    BIER \cite{bier_iccv} & 82.6 & 90.6 & -  & -  & 79.3 & 88.3 & -  & -  & 76.0 & 86.4 & -  & -  \\
    A-BIER \cite{a_bier}  & 86.3 & 92.7 & -  & -  & 83.3 & 88.7 & -  & -  & 81.9 & 88.7 & -  & -  \\
    FastAP~\cite{cakir2019fastAP} & 90.4 & 96.5 & -  & - & 88.0 & 95.4 & -  & -  & 84.5 & 93.9 & -  & -  \\
    \midrule
    DREML* \cite{dreml} & 88.5 & 94.8 & -  & -  & 87.2 & 94.2 & -  & -  & 83.1 & 92.4 & -  & -  \\
    \midrule
    Margin$\dagger$~\cite{margin} & 93.55 &  \textbf{97.09} & 90.62 & 90.53 & 91.77 & 96.04 & 90.56 & 88.65 & 89.75 & 95.26 & 90.03 & 86.22 \\

    \edited{Ours (fix) Not Progressive + Margin$\dagger$~\cite{sanakoyeu_dcesml}} & 94.81 & 96.37  & 89.89 & 93.83 & 93.88 & 95.67  & 90.41 & 92.98 & 93.50 & 95.78 & 90.54 & 92.34\\

    \textbf{Ours (fix) + Margin}$\dagger$ & \textbf{95.29} & \underline{96.78}  & 90.12 & \textbf{94.25} & \textbf{94.39} & \textbf{96.05}  & 90.59 & \textbf{93.37} & 93.96 & \textbf{96.24} & 90.78 & \textbf{92.82}\\


    \textbf{Ours (learn) + Margin }$\dagger$ & 95.10 & 96.67  & 90.04 & 93.98 & 94.18 & 95.97 & 90.56  & 93.21 & \textbf{94.03} & 96.18  & 90.67 & 92.73\\

    \bottomrule
    \end{tabular}
    }
\end{table*}

\begin{table*}[t]
    \rowcolors{2}{gray!8}{white}
    \centering
    \caption{Recall@k, NMI, and mARP on In-shop Clothes \cite{deepfashion}. $\dagger$ denotes our implementation with $d=512$; * denotes ensmebling methods or methods which require non-standard network architecture modifications.}
    \label{tab:inshop}

    \scalebox{0.9}{
    \begin{tabular}{lccccccc}
    \toprule
    R@k    & 1 & 10 & 20 & 30 & NMI & mARP\\
    \midrule
    FashionNet \cite{deepfashion} & 53.0 & 73.0 & 76.0 & 77.0 & - & -\\
    HDC \cite{hdc} & 62.1 & 84.9 & 89.0 & 91.2 & - & -\\
    HTG \cite{hard_triplet_gen} & 80.3 & 93.9 & 95.8 & 96.6 & - & -\\
    HTL \cite{htl} & 80.9 & 94.3 & 95.8 & 97.2 & - & -\\
    A-BIER \cite{a_bier} & 83.1	& 95.1 & 96.9 & 97.5 & - & -\\
    Stochastic Mining~\cite{suh2019softmax_and_triplet} & 88.9 & 97.2 & 98.2 & 98.6 & - & -\\
    NormSoftmax~\cite{zhai2018making_normsoftmax} & 88.6 & 97.5 & 98.4 & 98.8 & - & -\\
    MultiSim~\cite{wang2019multisim_loss} & 89.7 & 97.9 & 98.5 & 98.8 & - & - \\
    EPSHN~\cite{xuan2020easy_positive} & 87.8 & 95.7 & 96.8 & - & -  & -\\
    FastAP~\cite{cakir2019fastAP} & 90.0 & 97.5 & 98.3 & 98.7 & - & -\\
    \midrule
    ABE-8*~\cite{kim2018_abe} & 87.3 & 96.7 & 97.9 & 98.2 & - &-\\
    DREML* \cite{dreml} & 78.4 & 93.7 & 95.8 & 96.7 & - & -\\
    \midrule
    Margin$\dagger$ \cite{margin} & 87.87 &	96.26 & 97.27 & 97.83 & 89.16 & 86.14\\
    \edited{Ours (fix) Not Progressive + Margin$\dagger$~\cite{sanakoyeu_dcesml}} & 89.54 & 97.19 & 97.65 & 98.40 & 89.65 & 87.64\\
    \textbf{Ours (fix) + Margin$\dagger$} & 90.15 & \textbf{97.62} & \textbf{98.38} & \textbf{98.64} & \textbf{89.96} & 88.15\\
    \textbf{Ours (learn) + Margin$\dagger$} & \textbf{90.40} & 97.52 & 98.16 & 98.57 & 89.91 & \textbf{88.47}\\
    \bottomrule

    \end{tabular}}
\end{table*}

Tab.~\ref{tab:sota_results_cub_cars_sop}--\ref{tab:inshop} compare our proposed approach and the competitors on five benchmark datasets: CUB, CARS, Stanford Online Products, In-shop Clothes, and PKU VehicleID.
We can see that our approach significantly outperforms the state-of-the-art methods.
On CUB, CARS, and SOP datasets, our approach improves the R@1 by $1.2\%$, $2.6\%$, and $2.7$ over the baseline with margin loss \cite{margin}, achieving the state-of-the-art performance (see Tab.~\ref{tab:sota_results_cub_cars_sop}). On In-shop Clothes our approach improves over the margin loss baseline by $2.5\%$ in R@1 and outperforms the state-of-the-art methods like FastAP\cite{cakir2019fastAP}, ABE \cite{kim2018_abe} and MultiSim \cite{wang2019multisim_loss}. On the largest dataset, namely PKU VehicleID, our approach shows the greatest performance boost over the baseline -- it improves over the margin loss baseline by more than $4.2\%$ in R@1 on the large test leaving the runner-up approach FastAP\cite{cakir2019fastAP} almost $10\%$ behind.
FastAP\cite{cakir2019fastAP} utilizes a special hard-negative sampling technique which relies on the meta-class labels provided in the datasets, in contrast our approach does not use any extra labels and achieves implicit hard-negative sampling by the proposed progressive data division.
Our approach achieves especially high results compared to the previous state-of-the-art methods on large-scale datasets like Stanford Online Products, In-shop Clothes, and PKU VehicleID. Such a performance boost is explained by the hierarchical nature of our approach which gradually refines the learned metric by effective  decomposition of the large metric learning problem into sub-problems and distributing them to different subspaces of the embedding space.

Moreover, despite using a single Resnet-50 and only $512$ embedding dimensions, our approach outperforms even deep ensemble aporoaches like DREML \cite{dreml}, ABE-8 \cite{kim2018_abe}, and EDMS \cite{EDMS_2019_CVPR} on all presented datasets, proving the efficiency of our approach.
DREML \cite{dreml} utilizes $48$ ResNet-18 networks with the total number of $537$M trainable parameters.
In contrast, our model has only $26.6$M parameters and it outperforms DREML \cite{dreml} on CUB by $4\%$ and CARS by almost $2\%$ (see tab.~\ref{tab:sota_results_cub_cars_sop}). On the large-scale datasets In-shop Clothes \cite{deepfashion} and PKU VehicleID \cite{vehicleid} our method outperforms DREML by a large margin of at least $11\%$ in terms of Recall@1 (see Tab.~\ref{tab:inshop}, \ref{tab:vehicleid}).
EDMS \cite{EDMS_2019_CVPR} builds an ensemble using $25$ ResNet-18 networks with $128$ embedding dimensions each, which amounts to the massive final embedding of $3200$ dimensions. Nevertheless, our approach outperforms EDMS~\cite{EDMS_2019_CVPR} on all datasets (Tab.~\ref{tab:sota_results_cub_cars_sop}) despite our much more compact representation.


\section{Conclusion}
We have proposed a simple yet effective approach to address issues of existing metric learning by acting as a transparent wrapper that can be placed around arbitrary existing DML approaches.
Our approach gradually improves the quality of the learned embedding space by repeated simultaneous division of the data into more homogeneous sub-problems, which entail more valuable samples for training, and distributing them to different subspaces of the embedding space.
Learning several subspaces jointly, while each of them focuses on different latent  factors present in corresponding clusters, produces a more expressive embedding space with stronger generalization to novel classes.
The approach has been evaluated on five standard benchmark datasets for metric learning where it significantly improves upon the state-of-the-art.



\ifCLASSOPTIONcompsoc
  \section*{Acknowledgments}
\else
  \section*{Acknowledgment}
\fi

This work has been supported in part by the German Research Foundation (DFG) project 421703927 and the German Federal Ministry BMWi within the project ``KI Absicherung''.

\ifCLASSOPTIONcaptionsoff
  \newpage
\fi



\bibliographystyle{IEEEtran}
\bibliography{IEEEabrv,ref}

%

%
\input{biographies}

\appendices
\input{supplementary_text}

\end{document}

%% file: biographies.tex

\begin{IEEEbiography}[{\includegraphics[width=1in,height=1.25in,clip,keepaspectratio]{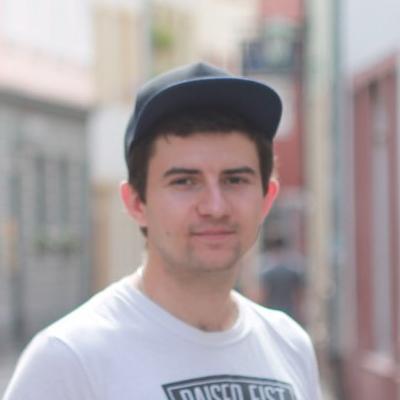}}]{Artsiom Sanakoyeu} defended his Ph.D. in Computer Vision at Heidelberg University. Currently he works as a Research Scientist at Facebook Reality Labs. His research interests include deep metric learning and representation learning for vision and image synthesis. Previously, he received a diploma in Computer Science from Belarusian State University, Minsk, Belarus. 
\end{IEEEbiography}


\begin{IEEEbiography}[{\includegraphics[width=1in,height=1.25in,clip,keepaspectratio]{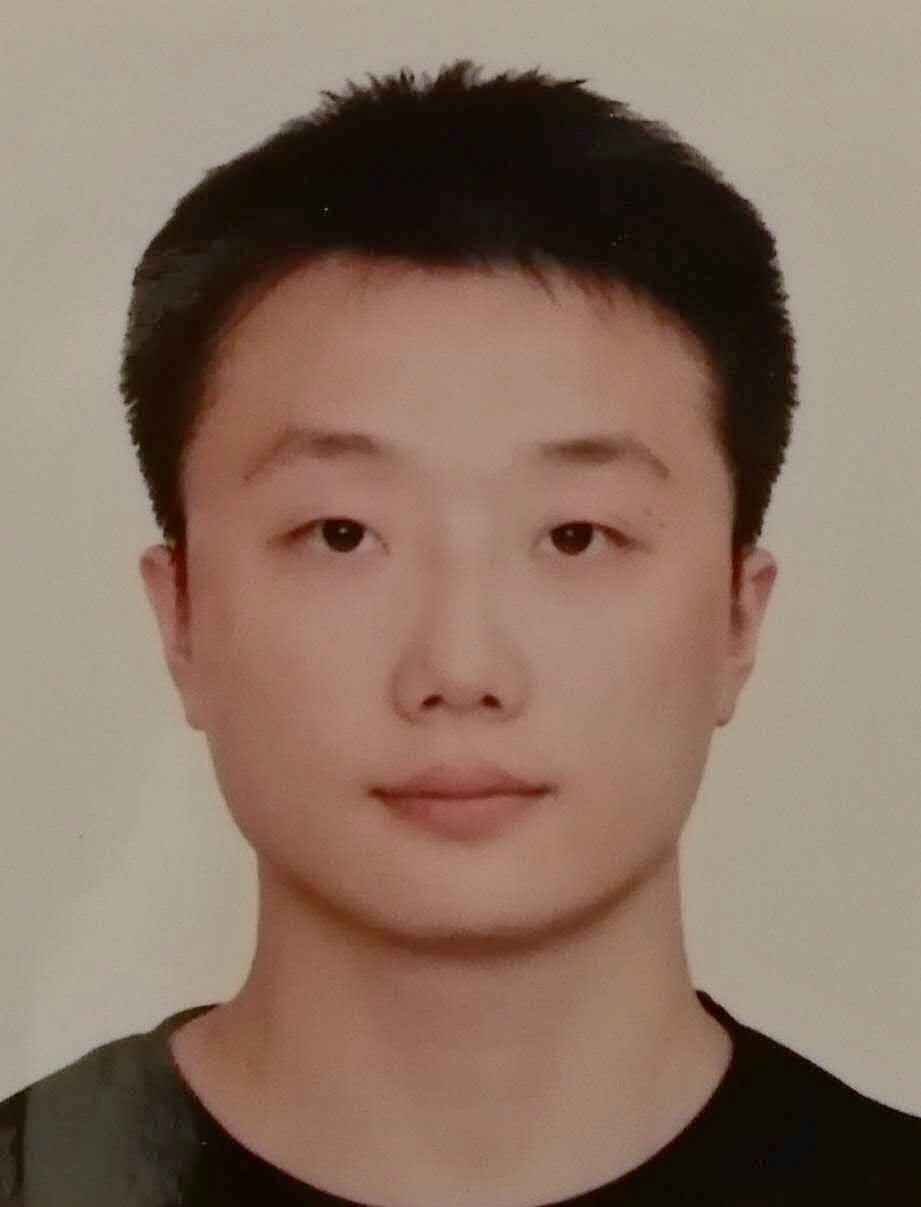}}]{Pingchuan Ma}
is a Ph.D. student in the Computer Vision research group at Heidelberg University. He is currently interested in topics such as deep metric learning, interpretable machine learning, and generative models. In 2015, he received his bachelor's degree in biomedical informatics from Guangzhou University of Chinese Medicine, Guangzhou, China.
\end{IEEEbiography}

\begin{IEEEbiography}[{\includegraphics[width=1in,height=1.25in,clip,keepaspectratio]{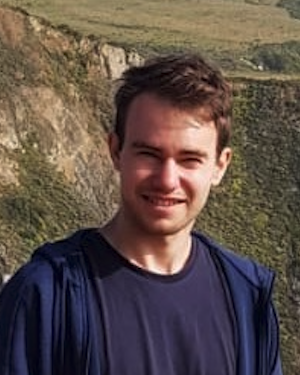}}]{Vadim Tschernezki} is a DPhil student in the Visual Geometry Group at the University of Oxford. Prior to that, he received his MSc degree in Applied Computer Science from Heidelberg University, Germany. His research focuses on computer vision and he is particularly interested in 3D reconstruction.
\end{IEEEbiography}

\begin{IEEEbiography}[{\includegraphics[width=1in,height=1.25in,clip,keepaspectratio]{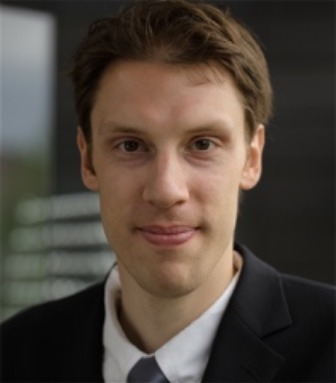}}]{Bj\"{o}rn Ommer}
received a diploma in computer science from the  University  of  Bonn,  Germany,   a  Ph.D.  from  ETH Zurich, and  held a postdoctoral position at the University of California at Berkeley. He then joint the Department of Mathematics and Computer Science at Heidelberg University as a professor where he is heading the computer vision group.  His research interests are in computer vision,  machine learning, and cognitive science. He is an associate editor of the IEEE Transactions on Pattern Analysis and Machine Intelligence.
\end{IEEEbiography}

%% file: supplementary_text.tex



\section{Datasets}
For evaluation we use five standard benchmark datasets commonly used in deep metric learning. 
Two smaller datasets: CARS196 \cite{cars196}, CUB200-2011 \cite{cub200_2011}; and three large-scale datasets: Stanford Online Products \cite{sop}, In-shop Clothes \cite{deepfashion}, and PKU VehicleID \cite{vehicleid}. For the retrieval and re-identification tasks we report Recall@k metric \cite{recallk}, and for measuring the clustering performance we compute the normalized mutual information score \cite{schutze2008introduction}
$$\text{NMI}(\Omega,\mathbb{C}) = \frac{2\cdot I(\Omega,\mathbb{C})}{H(\Omega)+H(\mathbb{C})},$$ where $\Omega$ is the ground truth clustering and $\mathbb{C}$ is the set of clusters obtained by K-means in the learned embedding space, $I$ denotes the mutual information and $H$ -- the entropy. However, the NMI score is not very informative because it depends on the clustering algorithm with most of algorithms not guaranteeing globally optimal solutions and depending on the random initialization. As an alternative we also report mean Average R-precision (mARP) which was proposed in \cite{musgrave2020}. Average R-Precision is equivalent to Average Precision at $k$ \cite{zhu2004recall_p_ap} where $k$ is the total number of relevant documents in the collection for the given query (i.e. $k$ would be different for different queries). mARP can be computed directly from the image embeddings without involving clustering and it rewards well clustered embedding spaces \cite{musgrave2020}.

\textbf{Stanford Online Products} (SOP) \cite{sop} is an image collection of products downloaded from eBay.com. For training, $11,318$ classes ($59,551$ images) out of $22,634$ classes are used, and $11,316$ classes ($60,502$ images) are held out for testing. We follow the same evaluation protocol as in \cite{lifted_struct}. 

\textbf{CARS196} \cite{cars196} contains 196 different types of cars distributed over 16,185 images. The first 98 classes ($8,054$ images) are used for training and the other 98 classes ($8,131$ images) for testing. We train and test on the entire images without using bounding box annotations.

\textbf{CUB200-2011} \cite{cub200_2011} consolidates images of 200 different bird species with 11,788 images in total and is an extended version of the CUB200 dataset. The first 100 classes ($5,864$ images) are used for training and the second 100 classes ($5,924$ images) for testing. We train and test on the entire images without using bounding box annotations.

\textbf{In-Shop Clothes Retrieval} \cite{deepfashion} contains $54,642$ images belonging to $11,735$ classes of clothing items. We follow the evaluation protocol of \cite{deepfashion} and use a subset of  $7,982$ classes with $52,712$ images. $3,997$ classes are used for training and $3,985$ classes for testing. The test set is partitioned into a query and gallery set, containing $14,218$ and $12,612$ images, respectively. 

\textbf{PKU VehicleID} (VID) \cite{vehicleid} is a large-scale vehicle dataset that contains $221,736$ images of $26,267$ vehicles captured by surveillance cameras. The training set contains $110,178$ images of $13,134$ vehicles and the testing set comprises $111,585$ images of $13,133$ vehicles. We follow the standard evaluation protocol of \cite{vehicleid} to evaluate the retrieval performance on $3$ test sets of different sizes. The small test set contains $7,332$ images of $800$ vehicles, the medium test set contains $12,995$ images of $1600$ vehicles, and the large test set contains $20,038$ images of $2400$ vehicles. This dataset has smaller intra-class variation, but it is more challenging than CARS196, because different identities of vehicles are considered as different classes, even if they share the same car model and color. 

\section{Hyperparameters}
In Tab.~\ref{tab:hyperparams} we show the detailed hyperparameter setting used in our experiments. The extra hyperparameters related to our hierarchical division have a limited range of values, and the optimal values are intuitively related to the size of a dataset. For example, maximum number of clusters $K_{max}$ can be from the set $\{2, 4, 8,16, 32, 64\}$, where $K_{max}=4$ is optimal for small datasets like CUB200-2011 and CARS196, and $K_{max}=8,16,32$ are optimal values for large datasets -- In-Shop Clothes, VID and SOP respecively. Regarding division frequency $E$, we explored the range $[1, 20]$, where $E=10$ was optimal for small datasets like CUB200-2011 and CARS196 and for all large datasets we set $E=2$.
We used cross-validation to find the optimal values.

\begin{table}[H]
    \centering
    \caption{Hyperparameters. $\infty$ denotes that there was no fine-tuning.}
    
    \scalebox{0.82}{
    \begin{tabular}{L|CC|CCC}
        \toprule
        Datasets $\rightarrow$ & CUB200-2011 & CARS196 & In-Shop & SOP  & VID \\
        \midrule
        Learning rate & \multicolumn{2}{c|}{$10^{-5}$} & 
        $10^{-5}$ & $10^{-5}$ & $10^{-5}$ \\
        Weigh decay & \multicolumn{2}{c|}{$10^{-3}$} & 
        $10^{-4}$ & $10^{-4}$ & $10^{-4}$ \\
        $\lambda$ &  \multicolumn{2}{c|}{$1.0$}& $0.2$ & $0.01$ & $5 \times 10^{-4}$ \\
        $K_{max}$ &  \multicolumn{2}{c|}{4} & 8 & 32 & 16 \\
        $E$ & \multicolumn{2}{c|}{10} & 2 & 2 & 2  \\
        Fine-tuning after epoch & \multicolumn{2}{c|}{$\infty$} & $\infty$ & 80 & 80 \\ 
        Total epochs & \multicolumn{2}{c|}{120} & 400 & 170 & 170\\
        \bottomrule
    \end{tabular}}
    \label{tab:hyperparams}
\end{table}

\section{Extra Experiments}
\subsection{Learned masks visualization}
In Fig.~\ref{fig:masks_heatmaps} we visualize the learned masks for the CUB200-2011 datasets, which became close to being mutually orthogonal after training. However, dimensions have continuous weights and the first mask has more dimensions with stronger weights than other masks, i.e. the embedding space is not equally partitioned in $4$ subspaces.
In Fig.~\ref{fig:masks_heatmaps_sop} we show masks learned on SOP dataset. In this case we have $K_{max} = 32$ sub-problems and we learned $32$ different masks. Some embedding dimensions are clearly shared across several subspaces because using orthogonal subspaces with $512/32=16$ dimensions each is not enough for learning informative representation.

\begin{table}[t]
    \centering
    \caption{Extra comparisons with local metric learning methods on CUB200-2011~\cite{cub200_2011}, CARS196~\cite{cars196} ($K_{max}=4$), and Stanford Online Products \cite{sop} ($K_{max}=32$). We report Recall@1 for every method.}
    
    \scalebox{0.95}{
    \begin{tabular}{L|CCC}
        \toprule
        Datasets $\rightarrow$ & CUB200-2011 & CARS196 & SOP  \\
        \midrule
        CLML~   \cite{saxena2015coordinated_LML} & 51.84 & 49.80 & 59.17 \\
        Ours & 68.16 & 87.76 & 79.77 \\
        \bottomrule
    \end{tabular}}
    \label{tab:extra_comparisons}
\end{table}

\subsection{UMAP visualization}
In addition to the plot in the main text, in Fig.~\ref{fig:umap_appendix} we display 2D projections of the learned image embeddings using UMAP \cite{umap} for the models trained on SOP with $K_{max} = 1,2,4,8,16,32$.
The baseline model ($K_{max} = 1$) overfits to the training classes (first column) and maps all the novel categories from the test set in few high-density hubs. 
However, when we increase the number of sub-tasks $K_{max}$ from $1$ to $32$ (Fig.~\ref{fig:umap_appendix}, rightmost column), both train and test embeddings become more uniformly spread over the embedding space and not crammed in few locations compared to the baseline. 
In Fig.~\ref{fig:average_emb_distance} we can see how expected distance from a query embedding to its nearest neighbors grows when we increase $K_{max}$ from $1$ to $32$, quantitatively demonstrating that embeddings become more uniformly distributed in the space.
This highlights the better utilization of the embedding space by our approach compared to the baseline, leading to less overfitting to the training data and improving generalization and overall performance.

\subsection{Comparison with local metric learning}
Existing local metric learning methods \cite{saxena2015coordinated_LML,amand2017sparse_comp_LML,bohne2018pairwise_LML} are not directly comparable with our approach, because they (a) learn \emph{linear} mappings for local regions of the data, while we learn a \emph{non-linear} embedding subspace for every data subset; (b) they did not evaluate on standard image retrieval benchmarks like CUB200-211~\cite{cub200_2011}, CARS196~\cite{cars196} or SOP~\cite{sop}; (c) used handcrafted features (not trained end-to-end); and (d) they used non-hierarchical data partitioning, while our approach is hierarchical -- we progressively increase the number of learned subspaces and corresponding data subsets. 

To make our comparisons more fair, we slightly modified Coordinated Local Metric Learning (CLML) \cite{saxena2015coordinated_LML} method. We use for CLML the same backbone (ResNet50~\cite{resnet} pretrained on ImageNet \cite{imagenet}), the same embedding size ($512$), and the same training settings as for our method (see Sect.~4.2 in the main manuscript). Moreover, we modified the CLML~\cite{saxena2015coordinated_LML} to backpropagate the gradients through the ResNet50 backbone in the same way as in our method and train with margin loss~\cite{margin}. Following \cite{saxena2015coordinated_LML}, the output of the backbone is first projected to $64$ dimensions using Principal Component Analysis (PCA) and then used for fitting a Gaussian Mixture Model with $K_{max}$ components. $K_{max} = 4$ for CUB200-211~\cite{cub200_2011} and CARS196~\cite{cars196} datasets, and $K_{max} = 32$ for SOP~\cite{sop} dataset.

In Tab.~\ref{tab:extra_comparisons} we show the comparison of the proposed approach with CLML~\cite{saxena2015coordinated_LML}. Our approach significantly outperforms CLML~\cite{saxena2015coordinated_LML} on three standard deep metric learning benchmarks (CUB200-211, CARS196 and SOP) confirming the superiority of the proposed model and importance of our design choices.


\begin{figure}[t]
\begin{center}
\includegraphics[width=\linewidth]{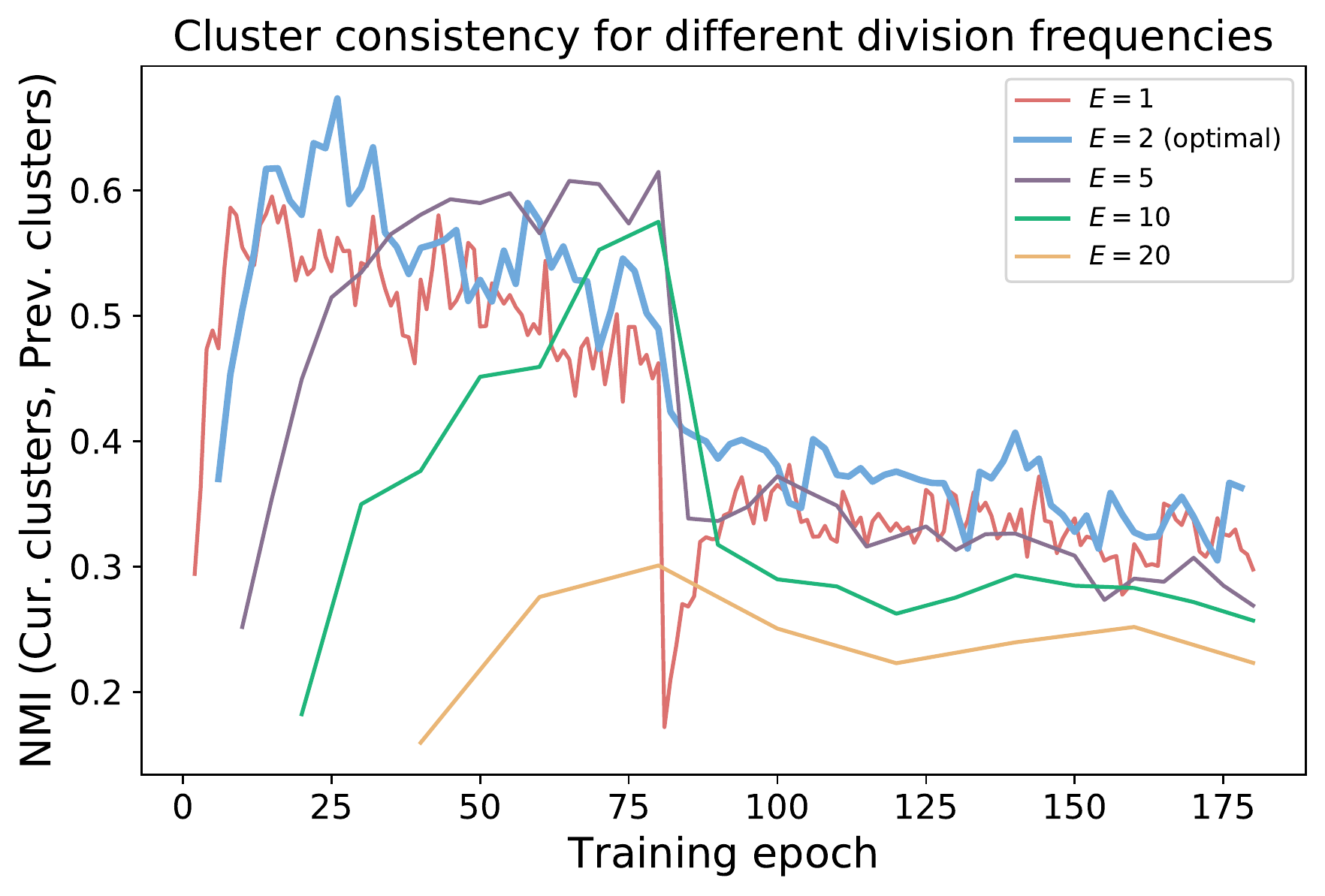} 

\end{center}
   \caption{\edited{NMI between clusters before and after each division step while training our model with margin loss and $K_{max}=32$ on SOP dataset. We vary the number of epochs $E$ between consecutive data division steps. $E=2$ corresponds to the model with the best retrieval performance.}}
\label{fig:cluster_consistency}
\end{figure}

\begin{figure*}[t!]
\begin{center}
  \includegraphics[width=0.9\linewidth]{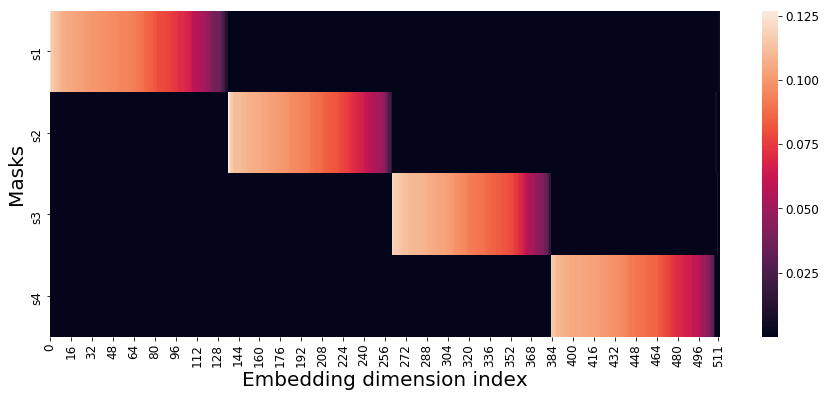}
\end{center}
  \caption{Visualization of the learned masks $s_1, s_2, s_3, s_4$ for CUB200-2011 dataset. We sorted the columns for the purpose of the visualization.}
\label{fig:masks_heatmaps}
\end{figure*}

\begin{figure*}[t!]
\begin{center}
  \includegraphics[width=0.9\linewidth]{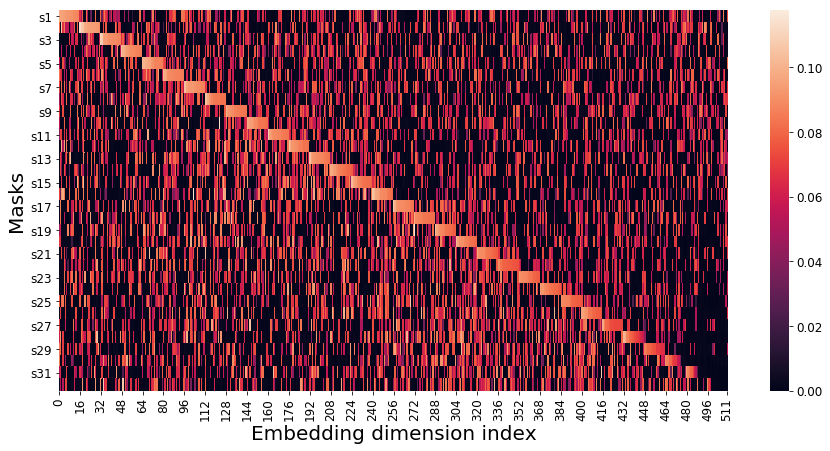}
\end{center}
  \caption{Visualization of the learned masks $s_1, s_2, \dots, s_{32}$ for SOP dataset. We sorted the columns for the purpose of the visualization.}
\label{fig:masks_heatmaps_sop}
\end{figure*}

\begin{figure*}[bt]
\begin{center}
\includegraphics[width=1.0\linewidth]{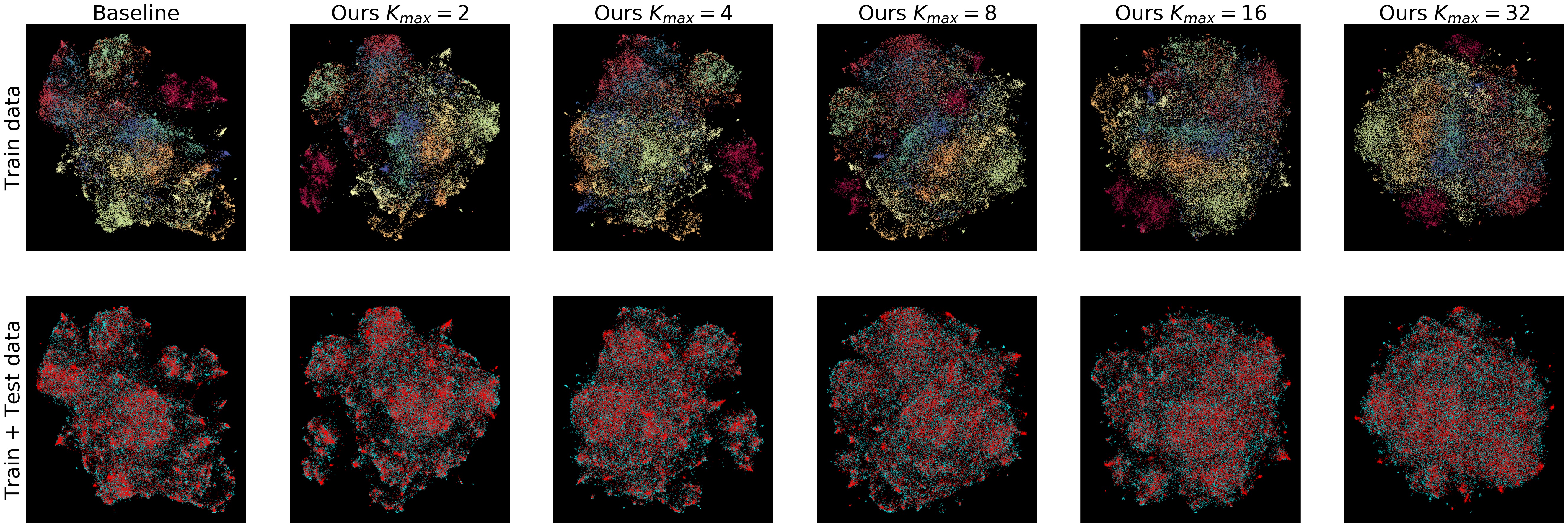}
\end{center}
  \caption{Low-dimensional projection of the embedding vectors produced by UMAP \cite{umap} for models trained on the SOP dataset with margin loss. We compare baseline ($K_{max}=1$) to our models with $K_{max}=4$ and $K_{max}=32$. \textbf{(1-st row)} we visualize embedding for training data, where color encodes class label ($11318$ classes); \textbf{(2-nd row)} embedding for \emph{train} (in \emph{blue}) and \emph{test} data (in \emph{red}) together. The embeddings produced by our model with $32$ clusters are more uniformly distributed in the space while still maintaining the superior retrieval performance.
  Moreover, we can see that compared to the baseline for $K_{max} = 32$ our test embedding distribution is less concentrated in few specialized areas and, thus, better matches the train distribution.
  This additionally underlines the better generalization of our model. Zoom-in for details.
  }
\label{fig:umap_appendix}
\end{figure*}

\subsection{\edited{Cluster consistency during training}}
\edited{
In Fig.~\ref{fig:cluster_consistency}, we demonstrate how we can implicitly control the degree of changes of cluster memberships by tuning the number of epochs $E$ between consecutive updates of the data partitioning. Similar to Fig.~4 (c) in the main text, here we plot NMI between cluster memberships of the samples before and after each division step.
As we can see, the highest on average cluster consistency is observed for $E=2$, which is also the value corresponding to the best performing model on SOP dataset (see Tab.3 in the main text). NMI for the case $E=2$ remains in the range of $[0.35, 0.6]$, which means that around $60\%-75\%$ of the points retain the same cluster label after reclustering and remapping using the linear assignment problem defined in Eq.~3. 
The resulting NMI score for the model with $E=2$ and its superior retrieval performance indicates that we get a sufficiently smooth transition between clusters.
}

\begin{figure*}[t!]
\begin{center}
\includegraphics[width=1.0\linewidth]{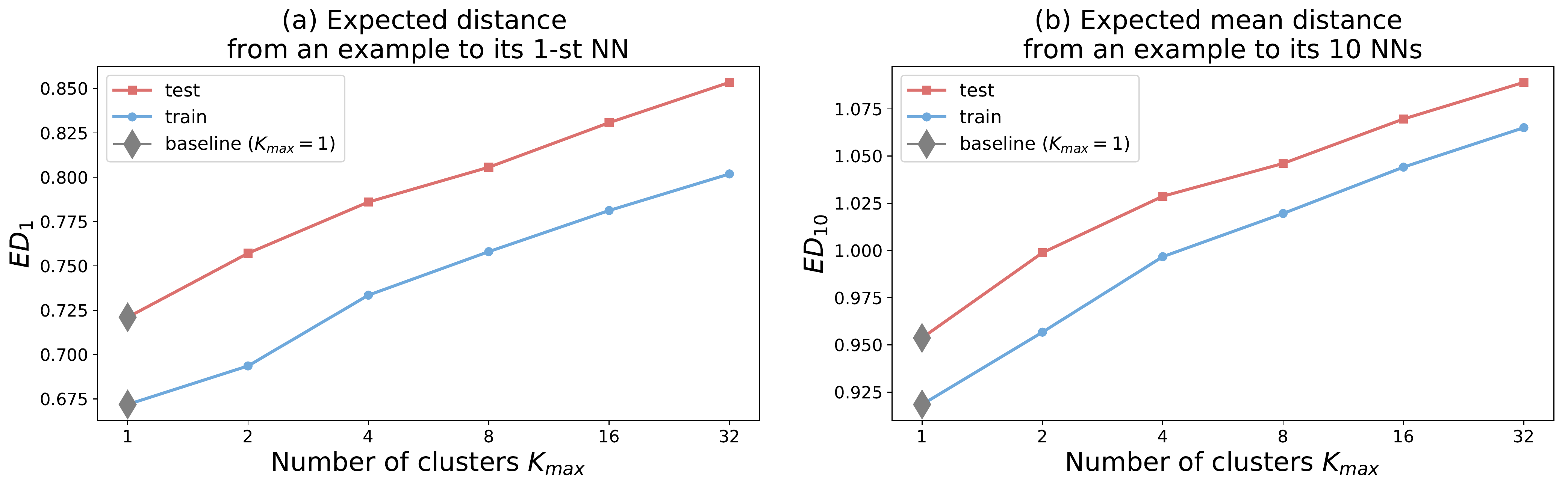}
\end{center}

  \caption{\textbf{(a)} Expected distance between an image embedding and its first nearest neighbor ($ED_1$); \textbf{(b)} expected mean distance from an image embedding to its ten nearest neighbors ($ED_{10}$) for our models with different number of sub-problems $K_{max}$ trained with margin loss \cite{margin} on SOP dataset. $K_{max} = 1$ is the baseline.
  The higher the expected distance, the more uniformly distributed the data in the embedding space. For reference, for uniformly distributed random vectors on a unit sphere,
  expected distance from a vector to its first NN is $1.275$ and expected average distance to its ten NNs is $1.290$.}
\label{fig:average_emb_distance}
\end{figure*}

